\documentclass{article}

\usepackage{microtype}
\usepackage{graphicx}
\usepackage{subfigure}
\usepackage[dvipsnames]{xcolor}
\usepackage{booktabs} 
\usepackage{nicefrac}
\usepackage{tablefootnote}

\usepackage{hyperref}

\usepackage[accepted]{icml2025}

\usepackage{amsfonts}
\usepackage{amssymb}
\usepackage{amsmath}
\usepackage{amsthm}
\usepackage{bm}
\usepackage{mathtools}
\mathtoolsset{showonlyrefs} 

\usepackage{enumitem}

\usepackage{mdframed}
\definecolor{theoremcolor}{rgb}{255, 255, 255}
\mdfsetup{
    backgroundcolor=theoremcolor,
    linewidth=1pt,
    leftline=true,
    topline=false,
    rightline=false,
    bottomline=false 
}

\newmdtheoremenv{definition}{Definition}
\newmdtheoremenv{proposition}{Proposition}
\newmdtheoremenv{corollary}{Corollary}
\newmdtheoremenv{theorem}{Theorem}
\newmdtheoremenv{lemma}{Lemma}
\newmdtheoremenv{example}{Example}

\def\e{{\bm{e}}}
\def\p{{\bm{p}}}
\def\q{{\bm{q}}}
\def\w{{\bm{w}}}
\def\x{{\bm{x}}}
\def\y{{\bm{y}}}

\def\thetav{{\bm{\theta}}}

\def\taumin{\tau_{\mathrm{min}}}
\def\taumax{\tau_{\mathrm{max}}}

\def\RR{\mathbb{R}}
\def\EE{\mathbb{E}}

\def\cW{\mathcal{W}}
\def\cX{\mathcal{X}}
\def\cY{\mathcal{Y}}

\def\ones{\bm{1}}

\def\softplus{\mathrm{softplus}}
\def\sigmoid{\mathrm{sigmoid}}
\def\softmax{\mathrm{softmax}}
\def\softargmax{\mathrm{softargmax}}

\DeclareMathOperator*{\argmax}{argmax}
\DeclareMathOperator*{\argmin}{argmin}
\DeclareMathOperator*{\dom}{dom}
\DeclareMathOperator*{\im}{im}

\def\aka{a.k.a.\ }
\def\wrt{w.r.t.\ }
\def\cf{c.f.\ }

\def\mytitle{Loss Functions and Operators Generated by $f$-Divergences}

\icmltitlerunning{\mytitle}

\begin{document}

\twocolumn[
\icmltitle{\mytitle}

\icmlsetsymbol{equal}{*}

\begin{icmlauthorlist}
\icmlauthor{Vincent Roulet}{gdm}
\icmlauthor{Tianlin Liu}{gdm}
\icmlauthor{Nino Vieillard}{gdm}
\icmlauthor{Micha\"{e}l E. Sander}{gdm}
\icmlauthor{Mathieu  Blondel}{gdm}

\end{icmlauthorlist}

\icmlaffiliation{gdm}{Google DeepMind}

\icmlcorrespondingauthor{Mathieu Blondel, Vincent Roulet}{mblondel@google.com, vroulet@google.com}

\icmlkeywords{loss functions, duality, $f$-divergences}

\vskip 0.3in
]



\printAffiliationsAndNotice{}  

\begin{abstract}
The logistic loss (\aka cross-entropy loss) is one of the most popular loss functions used for multiclass classification. It is also the loss function of choice for next-token prediction in language modeling. 
It is associated with the Kullback--Leibler (KL) divergence and the softargmax operator. 
In this work, we build upon Fenchel-Young losses to construct convex loss functions generated from $f$-divergences.  Our loss functions generalize the logistic loss in two directions: i) by replacing the KL divergence with $f$-divergences and ii) by allowing non-uniform reference measures. 
We instantiate our framework for numerous $f$-divergences, recovering existing losses and creating new ones.
By analogy with the logistic loss, the loss function generated by an $f$-divergence is associated with an operator, that we dub $f$-softargmax. We derive a novel parallelizable bisection algorithm for computing the $f$-softargmax associated with any $f$-divergence.
On the empirical side, one of the goals of this paper is to determine the effectiveness of loss functions beyond the classical cross-entropy in a language model setting, including on pre-training, post-training (SFT) and distillation. We show that the loss function generated by the $\alpha$-divergence (which is equivalent to Tsallis $\alpha$-negentropy in the case of unit reference measures) with $\alpha=1.5$ performs well across several tasks.
\end{abstract}

\section{Introduction}

The logistic loss, \aka cross-entropy loss, is widely used for multiclass classification. It is associated with the softargmax (\aka softmax) operator, which turns logits into class probabilities. The logistic loss and the softargmax operators are also frequently used in the space of tokens for language model pre-training, post-training and distillation. 

The classical softargmax is known to optimize a trade-off between the expected value of the logits and the Kullback--Leibler (KL) divergence with a uniform measure. This perspective is also adopted in the space of sequences, replacing the uniform measure with a reference measure, in reinforcement learning from human feedback (RLHF) \citep{christiano2017deep} and in direct preference optimization (DPO) \citep{rafailov2024direct}. Recent works extended this perspective to general $f$-divergences \citep{go2023aligning,wang2023beyond}.
However, these works naively compose an $f$-divergence with the (classical) softargmax, which does not result in a convex loss function.

In this work, building upon Fenchel--Young losses \citep{blondel2020learning},
we propose to construct new convex losses based on $f$-divergences.
Our losses generalize the logistic loss in two directions: i) by replacing the KL divergence with $f$-divergences and ii) by allowing non-uniform reference measures (class prior probabilities). 
Our loss construction generalizes the sparsemax \citep{martins2016softmax} and entmax \citep{entmax} losses, and allows us to create entirely new losses; see Table \ref{tab:loss_recap}. In addition, each loss generated by an $f$-divergence is associated with a new operator for turning logits into probabilities, that we dub $f$-softargmax.

On the empirical side, one of the goals of this paper is to determine the effectiveness of loss functions beyond the classical cross-entropy in a language model setting, including on pre-training, post-training (SFT) and distillation.

To summarize, we make the following contributions.
\begin{itemize}[topsep=0pt,itemsep=3pt,parsep=3pt,leftmargin=15pt]

\item We propose to use $f$-divergence regularization to generate loss functions. 
Each loss function is associated with a new operator, that we dub $f$-softargmax.
We instantiate this framework for numerous $f$-divergences, recovering existing losses and creating new ones.

\item We derive a novel bisection algorithm for computing the $f$-softargmax associated with any $f$-divergence. Our algorithm parallelizes well on modern hardware.

\item We demonstrate our loss functions on image classification, language model post-training and distillation.

\end{itemize}

\paragraph{Notation.}

Throughout this paper, $k$ denotes the number of classes in a classification problem or the vocabulary size in language modeling.
We denote the set $[k] \coloneqq \{1, \dots, k\}$.
We denote the probability simplex by $\triangle^k \coloneqq \{\p \in \RR_+^k \colon \langle \p, \ones \rangle = 1\}$.
We denote the domain of $f$ by $\dom(f) \coloneqq \{u \colon -\infty < f(u) < \infty\}$.
We denote the convex conjugate of $f$ by $f^*$, where $f^*(v) \coloneqq \sup_u uv - f(u)$.
We denote by $\thetav \coloneqq h_\w(\x)$ the logits produced by a network $h$ with parameters $\w \in \cW$ for the input $\x \in \cX$. We denote hard (one-hot) labels as $\y \in \{\e_1, \dots, \e_k\}$, which are used for classification. More generally, we denote soft labels as $\y \in \triangle^k$, which are useful for learning from label proportions, as in distillation.

\section{Background}

\subsection{Logistic loss}

Given logits $\thetav \in \RR^k$, we define the \textbf{softargmax} as
\begin{align}
[\softargmax(\thetav)]_j
& \coloneqq \frac{\exp(\theta_j)}{\sum_{j'=1}^k \exp(\theta_{j'})}
\quad j \in [k]. \label{eq:softargmax} 
\end{align}
Note that \eqref{eq:softargmax} is commonly known in the literature as ``softmax,'' which is a misnomer as it is a smooth approximation of the argmax function \citep{edpbook}. We instead define the \textbf{softmax} as
\begin{align}
\softmax(\thetav) & \coloneqq \log \sum_{j=1}^k \exp(\theta_j) \label{eq:softmax},
\end{align}
which is a smooth approximation of the maximum function. The softargmax and softmax are related by
\begin{equation}
\nabla \softmax = \softargmax.
\end{equation}

The logistic loss (a.k.a. cross-entropy loss) between logits $\thetav \in \RR^k$ and a ground-truth $\y \in \triangle^k$ is often defined as
\begin{equation}
\label{eq:cross-entropy} 
- \sum_{j=1}^k y_j \log \Big ( [\softargmax(\thetav)]_j \Big) 
= \softmax(\thetav) - \langle \thetav, \y \rangle.
\end{equation}
The logistic loss is equivalent up to a constant to the KL divergence between the target and the softargmax output,
\begin{align}
\ell(\thetav, \y) & \coloneqq \mathrm{KL}(\y, \softargmax(\thetav)) \\
& = \softmax(\thetav) - \langle \thetav, \y \rangle + \langle \y, \log \y \rangle \label{eq:logistic_kl}.
\end{align}
Adding the constant term $\langle \y, \log \y \rangle$ ensures that the loss is non-negative even when using soft labels (the constant term is zero with hard labels).
The loss $\ell$ is convex in $\thetav$, and its gradient \wrt $\thetav \in \RR^k$ is
\begin{equation}
\nabla_\thetav \ell(\thetav, \y) = \softargmax(\thetav) - \y.
\label{eq:logreg_grad}
\end{equation}

\subsection{Binary logistic loss}
\label{sec:binary_logistic_loss}

In the special case $k=2$, letting $\thetav \coloneqq (0, \theta)$ and $\y \coloneqq (1-y, y)$ for $\theta \in \RR$ and $y \in [0,1]$, we obtain the \textbf{softplus}
\begin{equation}
\softplus(\theta) \coloneqq \log(1 + \exp(\theta)) = \softmax(\thetav)
\end{equation}
and the \textbf{sigmoid}
\begin{equation}
\sigmoid(\theta) 
\coloneqq \frac{1}{1 + \exp(-\theta)}
= [\softargmax(\thetav)]_1.
\end{equation}
The two operators are again related by
\begin{equation}
\softplus' = \sigmoid.
\end{equation}
The binary logistic loss between the logit $\theta \in \RR$ and the ground-truth label 
$y \in [0,1]$ is then
\begin{equation}
\ell(\theta, y) 
\coloneqq \softplus(\theta) + (1-y)\log(1-y) + y \log y - \theta y.
\label{eq:binary_logistic}
\end{equation}
Beyond binary classification, a binary logistic loss can be used for pairwise ranking if we
define $\theta \coloneqq \theta_i - \theta_j$, where $\theta_i$ and $\theta_j$ are the scores of items $i \in [k]$ and $j \in [k]$, respectively. The probability that item $i \in [k]$ is ranked higher than item $j \in [k]$ according to the model is then 
\begin{equation}
\sigmoid(\theta_i - \theta_j) 
= \frac{\exp(\theta_i)}{\exp(\theta_i) + \exp(\theta_j)}.
\end{equation}
This is called the Bradley-Terry model (\citeyear{bradley1952rank}).

\subsection{Fenchel--Young losses}

It is well-known that the softmax and softargmax can be written from a variational perspective as
\begin{align}
\softmax(\thetav)
&= \max_{\p \in \triangle^k} \langle \p, \thetav \rangle - \langle \p, \log \p \rangle \\
\softargmax(\thetav)
&= \argmax_{\p \in \triangle^k} \langle \p, \thetav \rangle - \langle \p, \log \p \rangle.
\end{align}
This suggests that we can create more general softmax and softargmax operators if we replace
Shannon's negative entropy $\langle \p, \log \p \rangle$ by more general regularization $\Omega \colon \triangle^k \to \RR$, namely,
\begin{align}
\softmax_\Omega(\thetav)
&\coloneqq \max_{\p \in \triangle^k} \langle \p, \thetav \rangle - \Omega(\p) \\
\softargmax_\Omega(\thetav)
&\coloneqq \argmax_{\p \in \triangle^k} \langle \p, \thetav \rangle - \Omega(\p).
\end{align}
By analogy with \eqref{eq:logistic_kl}, we define the \textbf{Fenchel--Young} loss \citep{blondel2020learning} generated by $\Omega$ as
\begin{align}
\ell_\Omega(\thetav, \y) 
&\coloneqq \softmax_\Omega(\thetav)  - \langle \thetav, \y \rangle + \Omega(\y) \label{eq:fy_loss} \\
&\ge B_\Omega(\y, \softargmax_\Omega(\thetav)), \label{eq:composite_bregman}
\end{align}
where $B_\Omega$ is the Bregman divergence generated by $\Omega$.
The loss in \eqref{eq:fy_loss} is always convex \wrt $\thetav$, unlike the loss in \eqref{eq:composite_bregman}.
Among many other desirable properties, if $\Omega$ is strictly convex,
Fenchel--Young losses satisfy
\begin{equation}
\ell_\Omega(\thetav, \y) = 0 \iff \softargmax_\Omega(\thetav) = \y
\label{eq:identify_of_indiscernibles}
\end{equation}
and the gradient \wrt $\thetav \in \RR^k$ is
\begin{equation}
\nabla_\thetav \ell_\Omega(\thetav, \y) = \softargmax_\Omega(\thetav) - \y.
\end{equation}
In the binary classification setting, 
we can similarly replace $\sigmoid$ with $\sigmoid_\Omega$ in \eqref{eq:binary_logistic}, as well as the regularization term $(1-y)\log(1-y) + y \log y$ with $\Omega((1-y,y))$.

\section{Generating losses from $f$-divergences}
\label{sec:f_div_reg}

In this paper, we propose to study Fenchel--Young losses and associated operators when the regularizer is defined as
\begin{equation}
\Omega_f(\p; \q) \coloneqq D_f(\p, \q),
\label{eq:Omega_f_def}
\end{equation}
where $D_f$ is a $f$-divergence and $\q \in \RR_+^k$ is a reference measure, which contains the prior class weights. When $\q$ is not available, we can simply use $\q=\ones$, in which case we will obtain negative $f$-entropies, as explained below.

\begin{table}
\centering
\caption{By using $f$-divergences as regularization, 
we can generalize existing loss functions (logistic, sparsemax, entmax) to non-uniform reference measure $\q$
and we can construct several new loss functions.
Reverse $f$-divergences (not listed below), such as the reverse KL, can also be used to generate alternative loss functions. The $f$-softargmax operators associated with the Chi-square divergence and the $\alpha$-divergence for $\alpha > 1$ can produce probability distributions with sparse support. When using the unit positive measure $\q=\ones$, we obtain $f$-entropies, which recover existing known entropies (Shannon, Gini, Tsallis) up to a constant.
}
\begin{tabular}{cccc}
\toprule
Divergence & Entropy & Loss & Sparse \\
\midrule
Kullback--Leibler & Shannon & Logistic & No \\
Chi-square & Gini & Sparsemax & Yes \\
$\alpha$-divergence & Tsallis & Entmax & $\alpha > 1$ \\
Jensen--Shannon & --  & -- & No \\
Squared Hellinger & -- & -- & No \\
\bottomrule
\end{tabular}
\label{tab:loss_recap}
\end{table}

\subsection{$f$-divergences}


Let $f \colon \RR_+ \to \RR$ be a convex function such that 
$f(1) = 0$
and
$f(0)=\lim_{u\to 0^+} f(u)$.
The \textbf{$f$-divergence} between two discrete positive measures 
$\p \in \RR_+^k$ and $\q \in \RR_+^k$ 
\citep{renyi1961measures,csiszar1967information,ali1966general}
is then
\begin{equation}
D_f(\p, \q) 
\coloneqq \sum_{j=1}^k f(p_j / q_j)q_j
= \langle f(\p/\q), \q \rangle,
\end{equation}
where $f$ and division are applied element-wise.
An $f$-divergence is always non-negative and jointly convex in $\p$ and $\q$.
Many existing divergences can be written in $f$-divergence form: see Table \ref{tab:loss_recap} and Appendix \ref{app:f_div_examples}.

\paragraph{Divergence reversal.}

Since an $f$-divergence is not necessarily symmetric in $\p$ and $\q$, 
it may seem that setting the reference measure $\q$ as
right argument in \eqref{eq:Omega_f_def} is arbitrary.
However, if we define $g(u) \coloneqq u f(1/u)$, we obtain the reverse divergence of $D_f$,
$D_g(\p, \q) = D_f(\q, \p)$.
Therefore, for any $f$-divergence, we can always construct the corresponding reverse divergence, without loss of generality.

\subsection{$f$-entropies}
\label{sec:f_negentropies}

As we emphasized, our proposed framework can take into account a discrete reference measure $\q \in \RR_+^k$, which intuitively contains the prior class weights. When such a measure is not available, we can simply choose the unit positive measure $\q = \ones$.
In this case, we recover negative $f$-entropies \cite{cichocki2010families}, \aka \textbf{$f$-negentropies}. They are defined as
\begin{equation}
\Omega_f(\p) 
\coloneqq \sum_{j=1}^k f(p_j)
= \langle f(\p), \ones \rangle
= D_f(\p, \ones).
\end{equation}
By choosing $f$, we recover (up to a constant) numerous existing negentropies.
When $f(u) = u \log u$, which is the generating function of the KL divergence,
we recover the \textbf{Shannon negentropy},
\begin{equation}
\Omega_f(\p) = \langle \p, \log \p \rangle.
\end{equation}
When $f(u) = \frac{1}{2} (u^2 - 1)$, which is the generating function of the Chi-square divergence, we recover the \textbf{Gini negentropy}, used for the sparsemax loss \cite{martins2016softmax},
\begin{equation}
\Omega_f(\p) \doteq \frac{1}{2}(\|\p\|_2^2 - 1)
\end{equation}
(we use $\doteq$ for equality up to a constant).
More generally,
with $f(u) = \frac{(u^\alpha - 1) - \alpha(u - 1)}{\alpha(\alpha-1)}$, 
which is the generating function of $\alpha$-divergences (using \citet{cichocki2010families}'s definition),
we recover the \textbf{Tsallis negentropy}, 
\begin{equation}
\Omega_f(\p) 
\doteq \frac{1}{\alpha} \langle \p, \log_\alpha(\p) \rangle
= \frac{1}{\alpha(\alpha-1)}(\|\p\|_\alpha^\alpha - 1),
\end{equation}
used by the entmax loss \citep{entmax}.
The Tsallis negentropy itself is very general, as it recovers the Shannon negentropy when $\alpha \to 1$ and the Gini negentropy when $\alpha = 2$. In our experiments, we will demonstrate good results with the choice $\alpha=1.5$, which can be thought as a middle ground between Shannon entropy (used by the logistic loss) and Gini entropy (used by the sparsemax loss).

\paragraph{Unexplored entropy instances.}

\begin{figure}[t]
    \centering
    \includegraphics[scale=0.42]{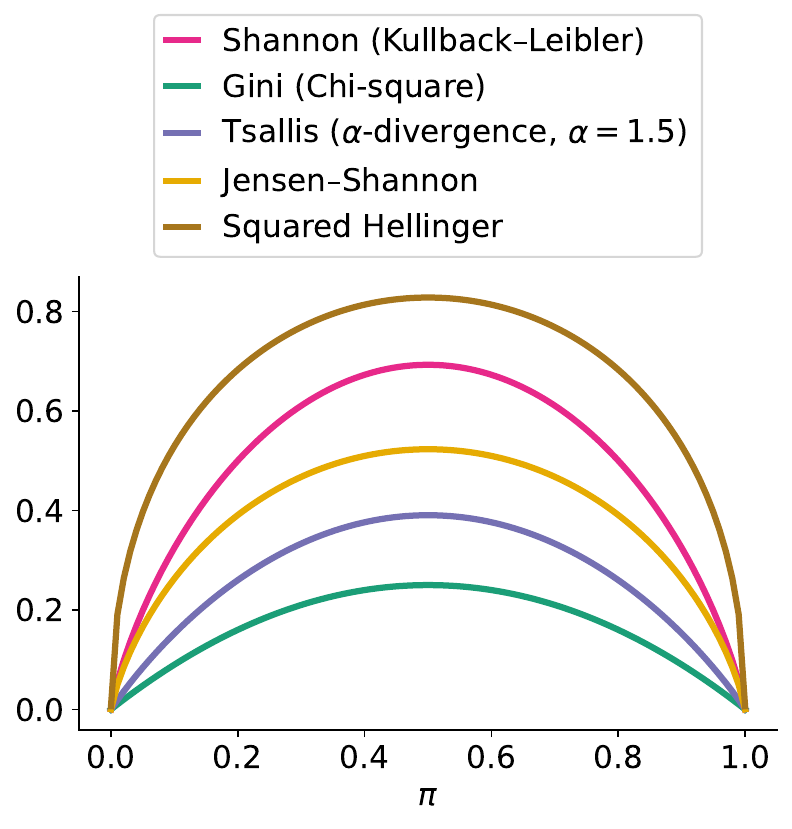}
    \caption{Illustration of $f$-entropies $-\Omega_f(\p)$ for $\p = (1-\pi,\pi)$ and varying $\pi \in [0,1]$.
    We add a constant $f(0)$ to ensure non-negativity of the $f$-entropies.
    }
    \label{fig:entropies}
\end{figure}

Beyond existing entropies, some choices of $f$ lead to entropies that had not been considered before, such as the squared Hellinger, illustrated in Figure~\ref{fig:entropies}. We leave a formal  axiomatic analysis of such new entropies to future work, and refer the interested reader to, e.g., \citet{ben1978f, friedland2013universal,sharma2021geometric, sbert2024entropies}.

\paragraph{Effective domain.}

We point out, however, that some $f$-entropies are only well-defined on the relative interior of the probability simplex, if $\lim_{u \to 0} f(u) = -\infty$. This is for instance the case of the $f$-entropies associated with the reverse KL and Jeffrey divergences.
This means that the loss functions generated by these choices only work with strictly positive soft labels.

\subsection{$f$-softmax and $f$-softargmax}

Overloading the notation,
we define the \textbf{$f$-softmax} as
\begin{equation}
\softmax_f(\thetav; \q)
\coloneqq \max_{\p \in \triangle^k} \langle \p, \thetav \rangle - D_f(\p, \q) \in \RR
\label{eq:f_softmax}
\end{equation}
and the \textbf{$f$-softargmax} as
\begin{equation}
\softargmax_f(\thetav; \q)
\coloneqq \argmax_{\p \in \triangle^k} \langle \p, \thetav \rangle - D_f(\p, \q) \in \triangle^k.
\label{eq:f_softargmax}
\end{equation}
Compared to a classical softmax and softargmax,
our operators use a function $f$ and include an additional reference measure $\q$ as argument. When $f(u) = u \log u$ and $\q=\ones$, we recover the classical softmax and softargmax.

Scaling the divergence by a temperature parameter $\beta > 0$ can easily be done. 
Indeed, for any $\beta >0$,
\begin{align*}
\softmax_{\beta f}(\thetav; \q) &= \beta \softmax_f(\thetav/\beta; \q) \\
\softargmax_{\beta f}(\thetav; \q) &= \softargmax_f(\thetav/\beta; \q).
\end{align*}

\begin{figure}[t]
    \centering
    \includegraphics[width=\linewidth]{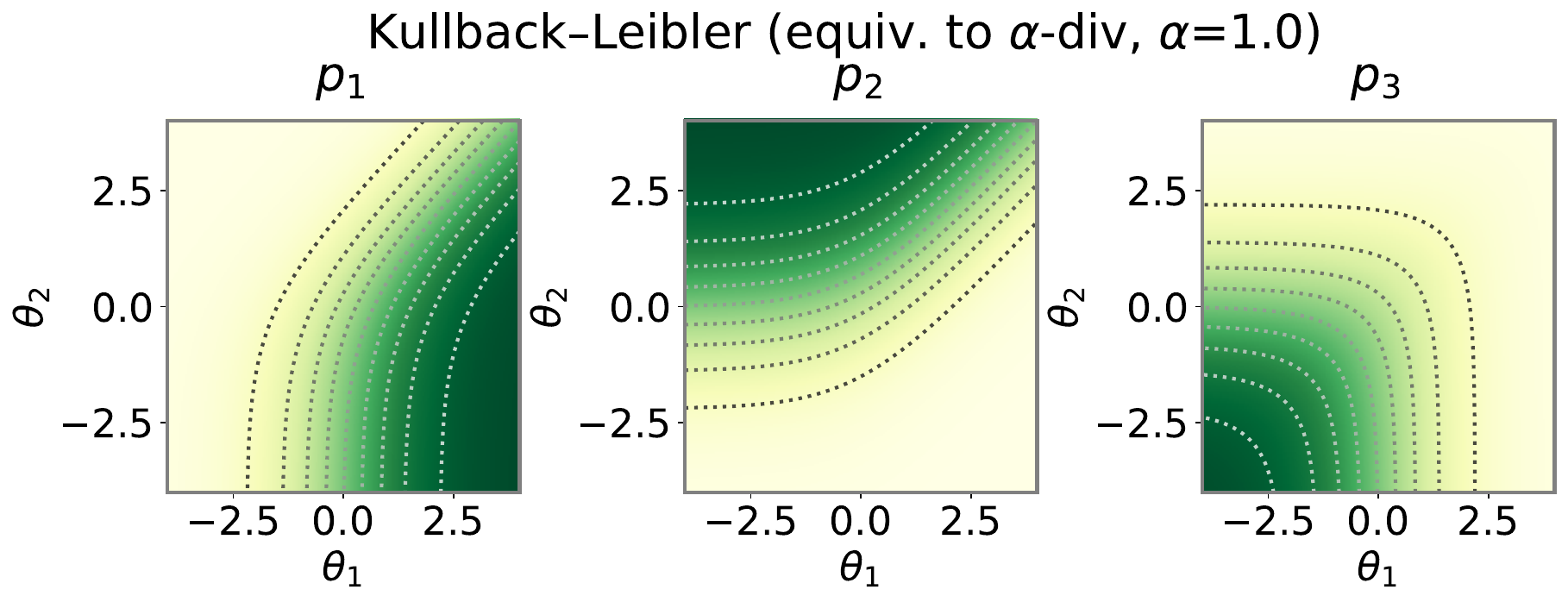}
    \includegraphics[width=\linewidth]{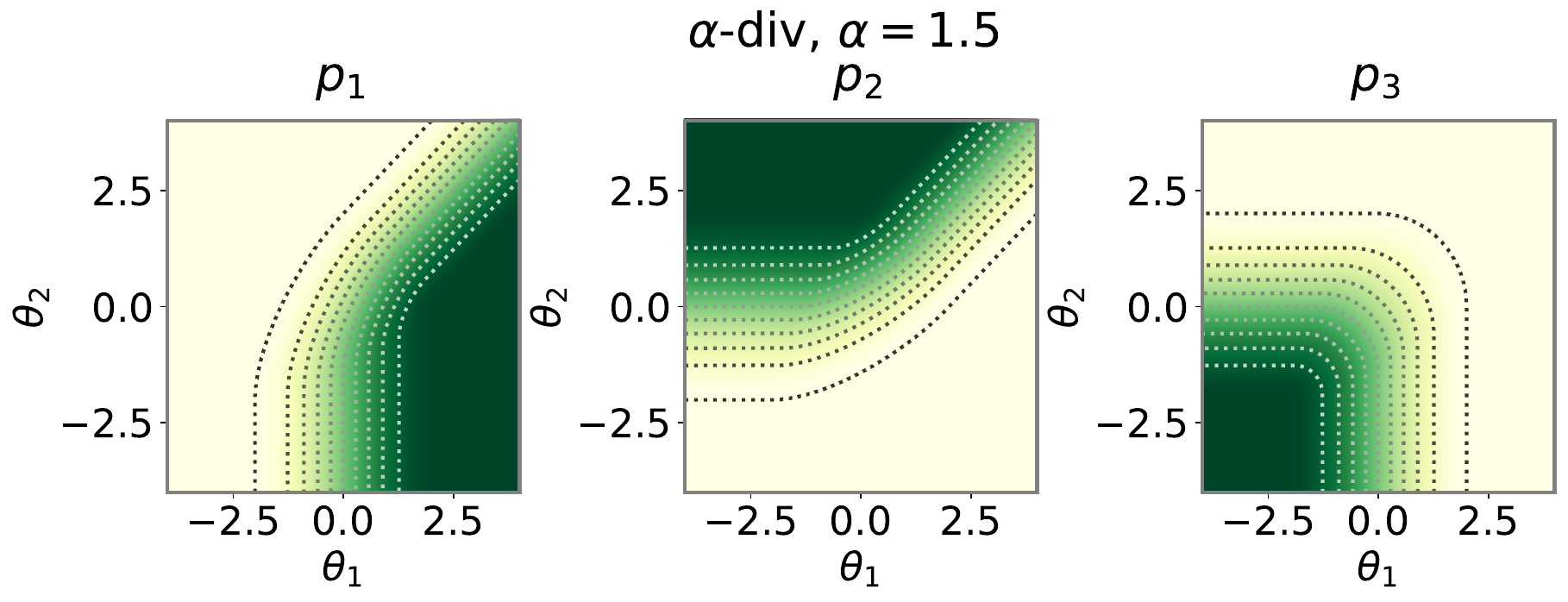}
    \includegraphics[width=\linewidth]{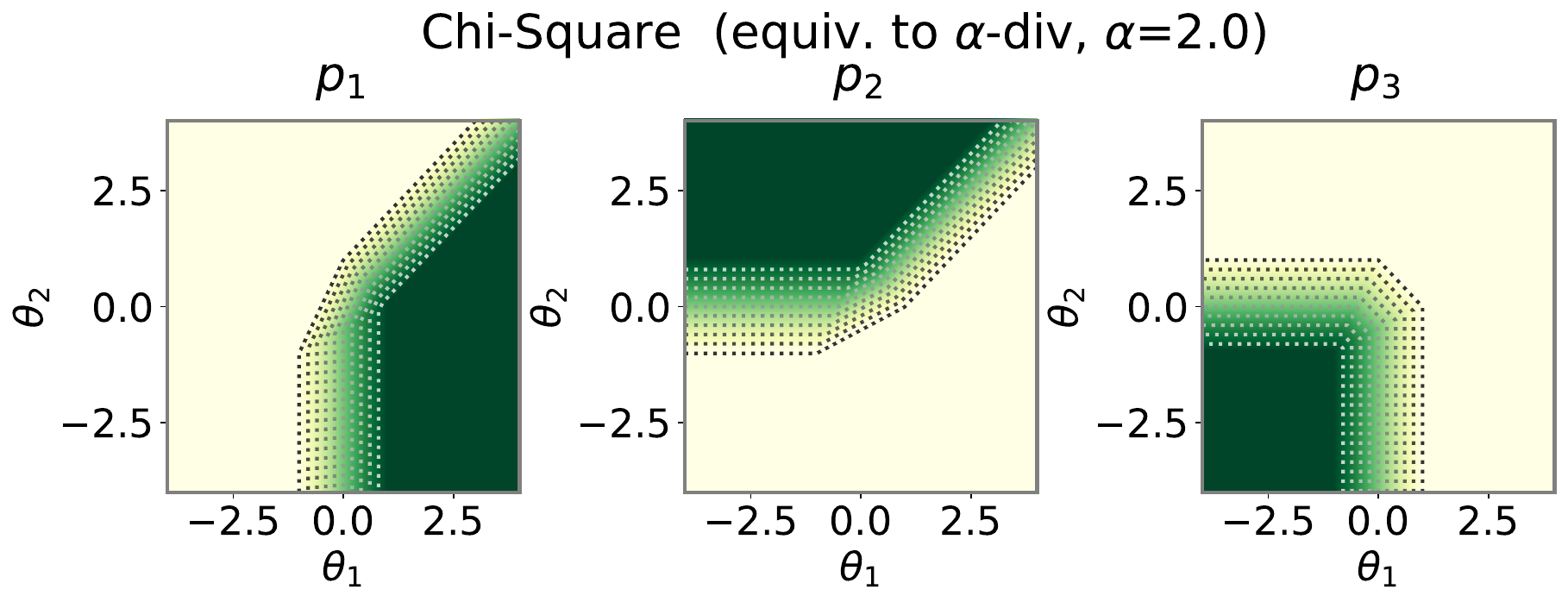}
    \hspace*{20pt}\includegraphics[width=0.4\linewidth]{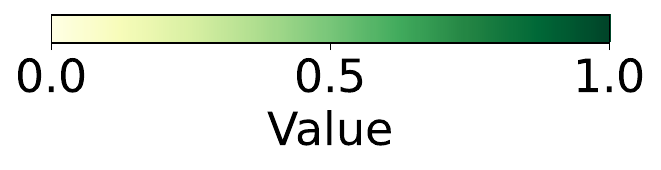}
    \caption{Illustration of $(p_1, p_2, p_3) = \softargmax_f(\theta_1, \theta_2, 0; \q)$ when varying $\theta_1, \theta_2 \in \RR$ for three possible divergences and with $\q = (1, 1, 1)$. More illustrations are given in Appendix \ref{app:f_div_examples}.}
    \label{fig:softargmax_main_plots}
\end{figure}

\paragraph{Sparse distributions.}

As summarized in Table \ref{tab:loss_recap},
the $f$-softargmax associated with the Chi-square and $\alpha$ divergences for $\alpha > 1$
can produce probability distributions with sparse support, meaning that some classes have exactly zero probability according to the model. As will be clear from Proposition \ref{prop:computation}, the $f$-softargmax can be sparse when
$0 \in \dom(f') \iff \lim_{u\rightarrow 0} f'(u) > -\infty$.

\subsection{$f$-softplus and $f$-sigmoid}

As special cases of $f$-softmax and $f$-softargmax, 
by defining
$\thetav \coloneqq (0, \theta)$, $\p \coloneqq (1-\pi, \pi)$ and $\q \coloneqq (q_0, q_1)$,
we obtain the \textbf{$f$-softplus}
\begin{equation}
\softplus_f(\theta; \q) 
\coloneqq \max_{\pi \in [0,1]} \pi \theta - D_f((1-\pi, \pi), \q) \in \RR
\end{equation}
and the \textbf{$f$-sigmoid}
\begin{equation}
\sigmoid_f(\theta; \q)
\coloneqq \argmax_{\pi \in [0,1]} \pi \theta - D_f((1-\pi, \pi), \q) \in [0,1].
\end{equation}
The effect of the prior class weights $\q \in \RR_+^2$ on the shape of the sigmoid is illustrated in Figure \ref{fig:sigmoid} in Appendix~\ref{app:f_div_examples}.
The $f$-softplus and $f$-sigmoid can be used, not only for binary classification, but also for pairwise ranking, by analogy with Section \ref{sec:binary_logistic_loss}.
Similarly to the $f$-softmax and $f$-softargmax, we can easily scale $D_f$ by a temperature parameter $\beta > 0$.


\subsection{Loss function}

To obtain a loss function associated with the $f$-softargmax, we instantiate the Fenchel--Young loss defined in \eqref{eq:fy_loss} with the regularization defined in \eqref{eq:Omega_f_def} to define
\begin{equation} \label{eq:f-div-loss}
\begin{aligned}
\ell_f(\thetav, \y; \q) 
&\coloneqq \softmax_f(\thetav; \q) + \Omega_f(\y; \q) - \langle \thetav, \y \rangle \\
&= \softmax_f(\thetav; \q) + D_f(\y, \q) - \langle \thetav, \y \rangle,
\end{aligned}
\end{equation}
where $\thetav \in \RR^k$ are the logits, $\y \in \triangle^k$ is the ground-truth and $\q \in \RR_+^k$ is a reference measure, which contains the prior class weights. This loss inherits from all the desirable properties of Fenchel--Young losses. In particular, it is convex \wrt $\thetav$ and it is differentiable everywhere if $f$ is strictly convex. On the other hand, it is not necessarily convex \wrt $\q$, since $\softmax_f$ involves the maximum over a collection of \textit{concave} functions of $\q$. Because $\y \in \triangle^k$ and not just $\y \in \{\e_1,\dots,\e_k\}$, this loss can also be used for learning from label proportions, as is useful in distillation for example.
\vspace{-1em}
\paragraph{Choosing $\q$.}

One distinctive feature of our losses compared to usual instances of Fenchel--Young losses is the possibility to adjust the reference measure $\q$. 
In the absence of prior knowledge, we can simply use $\q = \ones$ or $\q = \ones / k$. This recovers Fenchel--Young losses generated by $f$-negentropies.
If class weights are available, we can use this prior knowledge as $\q$.

\subsection{Computation}

On first sight, it is not obvious how to solve the variational problems \eqref{eq:f_softmax} and \eqref{eq:f_softargmax} involved in computing the $f$-softmax and $f$-softargmax, respectively. We need an algorithm that works for any valid choice of $f$ and parallelizes well, as we often need to compute the operators on a batch of $b$ logits $\thetav_{i_1}, \dots, \thetav_{i_b}$, where $\thetav_{i_j} \coloneqq h_\w(\x_{i_j})$.

In this section, we introduce a novel generic algorithm for computing the $f$-softmax and $f$-softargmax given access to $f$ and $f^*$, where $f^*$ is the convex conjugate of $f$.
The latter is usually available in closed form; we give numerous examples in Appendix \ref{app:f_div_examples}.

For convenience, with a slight abuse of notation, we define the shorthand $f_*' \coloneqq (f^*)'$. We denote $f'(0) = \lim_{x\rightarrow 0, x\geq 0} f'(x)$ which may be finite or $-\infty$.
\begin{proposition}{Reduction to root finding}
\label{prop:computation}

Let $f$ be a strictly convex and differentiable function such that $(0, +\infty) \subseteq \dom f'$.
Then, for any $\thetav \in \RR^k$,
\vspace{-1.em}
\begin{align}
&\softmax_f(\thetav; \q) = \tau^\star + \sum_{j=1}^k q_j f^*(\max\{\theta_j - \tau^\star, f'(0)\}) \label{eq:f_softmax_sol} \\
&[\softargmax_f(\thetav; \q)]_j = q_j f_*'(\max\{\theta_j - \tau^\star, f'(0)\}),
\label{eq:f_softargmax_sol}
\end{align}
where $\tau^\star$ is the unique solution of
\begin{equation}
\label{eq:root_equation}
\sum_{j=1}^k q_j f_*'(\max\{\theta_j - \tau, f'(0)\}) = 1,
\end{equation}
on $\tau \in [\tau_{\min}, \tau_{\max}]$, where, for $j^\star \in \argmax_{j\in [k]} \theta_j$,
\vspace{-1em}
\begin{align*}
\tau_{\min} & \coloneqq \theta_{j^\star} - f'(1/q_{j^\star}) \\
\tau_{\max} & \coloneqq \theta_{j^\star} - f'\left(1/\left(\textstyle{\sum_{j=1}^k} q_j\right)\right).
\end{align*}
\end{proposition}
A proof is given in Appendix \ref{proof:computation}.
Proposition \ref{prop:computation} is a generalization of \citep[Proposition 9]{blondel2020learning} to arbitrary reference measures $\q$.
Our proof technique is different: it uses Fenchel duality as opposed to Lagrange duality and it rigorously accounts for the domain of $f'$, which is one of the technical difficulties for supporting general $f$-divergences.
Proposition \ref{prop:computation} is a generalization of \citep[Theorem 1]{wang2023beyond} to the case $0 \in \dom(f')$, that is, to $f$-softargmax operators with sparse output. 
A Newton algorithm was proposed in \citep{terjek2021moreau} for computing the same operator. However, that algorithm was used for regularized optimal transport, not for creating loss functions. In addition, it assumes that $(f + \delta_{\RR_+})^*$ is twice differentiable, which is not the case when $0 \in \dom(f')$, and does not enjoy convergence guarantees, as a line search was not used.
See also the discussion in \citep{belousov2017f} on the implications of $0 \not \in \dom(f')$.

\paragraph{Implementation.}
In practice, we solve the root equation in \eqref{eq:root_equation} by bisection (Algorithm~\ref{algo:bisect}). This algorithm has several advantages: it is simple, parallelizes well on GPU and TPU (we often need to compute the $f$-softargmax of a batch) and achieves an error on iteration $t$ of $(\taumax - \taumin)/2^t$.
That is, the error exponentially decreases with the number of iterations. We show in Appendix \ref{app:comp_cost} that the overhead of our proposed algorithm is negligible compared to a classical softargmax.

\begin{algorithm}[H]
\centering
    \caption{Computing $f$-softmax and $f$-softargmax}
\begin{algorithmic}
    \STATE \textbf{Inputs:} logits $\thetav \in \RR^k$, prior $\q \in \RR_+^k$, tolerance $\epsilon > 0$
\STATE $[\p(\tau)]_j \coloneqq q_j f_*'(\max\{\theta_j - \tau, f'(0)\})$, $j \in [k]$
\STATE $\phi(\tau) \coloneqq \langle \p(\tau), \ones \rangle - 1$
\STATE $j^\star \in \argmax_{j\in [k]} \theta_j$
\STATE $\tau_{\min} \leftarrow \theta_{j^\star} - f'(1/q_{j^\star})$
\STATE $\tau_{\max} \leftarrow \theta_{j^\star} - f'\left(1/\left(\textstyle{\sum_{j=1}^k} q_j\right)\right)$
\STATE  $\tau \leftarrow (\tau_{\min} + \tau_{\max}) / 2$
\STATE \textbf{while} $|\phi(\tau)| > \epsilon$
\STATE \hspace{0.4cm}\textbf{if} $\phi(\tau) < 0$
\STATE \hspace{0.8cm}$\tau_{\max}\leftarrow\tau$ 
\STATE \hspace{0.4cm}\textbf{else}
\STATE \hspace{0.8cm} $\tau_{\min}\leftarrow\tau$ 
\STATE \hspace{0.35cm} $\tau \leftarrow (\tau_{\min} + \tau_{\max}) / 2$
\STATE \textbf{Outputs:} $\softargmax_f(\thetav; \q) \approx \p(\tau)$
\STATE $\softmax_f(\thetav; \q) \approx \tau + \sum_{j=1}^k q_j f^*(\max\{\theta_j - \tau, f'(0)\})$
\end{algorithmic}
\label{algo:bisect}
\end{algorithm}

\paragraph{Closed forms in the binary case.}

In the special case $k=2$ (binary classification), we can often derive closed-form solutions for the $f$-softplus and the $f$-sigmoid operator. For completeness, we derive the expressions for numerous cases in Appendix \ref{app:f_sigmoids}.

\subsection{Differentiation}

\paragraph{Differentiating through $f$-softmax and loss.}

In order to differentiate the $f$-softmax, we can simply use Danskin's theorem to obtain
\begin{align}
\nabla_\thetav \mathrm{softmax}_f(\thetav; \q) &= \p^\star \coloneqq \softargmax_f(\thetav; \q)\\
\nabla_\q \mathrm{softmax}_f(\thetav; \q) &= -\nabla_\q D_f(\p^\star, \q).
\end{align}
As a result, the loss gradients are
\begin{align}
\nabla_\thetav \ell_f(\thetav, \y; \q) &= \p^\star - \y = \softargmax_f(\thetav; \q) - \y \\
\nabla_\q \ell_f(\thetav, \y; \q) &= \nabla_\q D_f(\y, \q) -  \nabla_\q D_f(\p^\star , \q).
\end{align}
The first equation is in complete analogy with \eqref{eq:logreg_grad}.
The second equation is the difference of the $f$-divergence gradients evaluated at the ground-truth $\y$ and the prediction $\p^\star$.
Differentiating through the $f$-softplus operator and its associated loss function is similar.

\paragraph{Differentiating through $f$-softargmax.}

When the goal is to use the $f$-softargmax as the operator associated with our loss functions,
we do not need to differentiate through the $f$-softargmax. As explained above,
thanks to Danskin's theorem, differentiating through the $f$-softmax is sufficient. 
When the goal is to use the $f$-softargmax as an attention mechanism, however,
we do need to differentiate through the $f$-softargmax. This is more challenging as,
from Proposition \ref{prop:computation},
we need to differentiate through the solution of a root equation.
Under assumptions on $f$, we can apply the implicit function theorem \citep{krantz2002implicit} through the root equation's solution. This can be implemented using automatic implicit differentiation \citep{blondel_implicit_diff}. Importantly, implicit differentiation does not require solving a costly linear system here. Since the root equation is one-dimensional, a simple division is sufficient.


\section{Experiments}

To evaluate different $f$-divergence generated losses, we apply them to tasks of different data modalities, including image classification (Section~\ref{subsec:imagenet}) and text generation (Section~\ref{subsec:lm}). 
These experiments also cover different training strategies: from scratch, finetuning, and distillation. 



\subsection{ImageNet classification \label{subsec:imagenet}}

We apply different $f$-divergence generated losses to train a ResNet50 model \citep{he2016deep} on the ImageNet-2012 dataset \citep{russakovsky2015imagenet}. The ImageNet dataset contains 1.28 million training images and 50,000 validation images, belonging to one of 1,000 classes. The ResNet50 model is a standard choice for ImageNet.

We use an SGD optimizer with 0.9 momentum to train the ResNet50 model for 90 epochs. During the initial 5 epochs, we use a linear warmup that achieves a peak learning rate of 0.2; we then use cosine annealing to reduce the learning rate to 0. The weight decay is set to be $10^{-4}$. The batch size is 512.

Table~\ref{tab:acc_imagenet} shows the validation accuracy of training ResNet50 using different $f$-divergence generated losses; boldface indicates the highest accuracy. Recall that the KL generated loss is equivalent to the standard cross-entropy loss; therefore, using the KL generated loss should match the result of prior work \citep{he2016deep}. Our experiments confirm this: the KL loss reaches 76.87\% accuracy, consistent with previous benchmarks (Appendix~\ref{appendix:imagenet-classification}).  Perhaps surprisingly, we find that the $\alpha$-divergence loss function surpasses the KL loss in validation accuracy. This improvement is achieved without adjusting any hyperparameters and therefore is notable. In addition, $\alpha = 1.5$ seems to be optimal within $\alpha$-divergences as shown in Figure \ref{fig:app_imagenet}, where we see that validation accuracy is maximal for $\alpha = 1.5$ among $11$ values for $\alpha \in [1, 2]$.
However, the Jensen–Shannon, Squared Hellinger, and Chi-square (Pearson divergence) generated loss functions perform worse than the KL divergence in this experiment.

\begin{table}[t]
\centering
\caption{ImageNet classification results.}
\begin{tabular}{lc}
\toprule
Divergence & Accuracy (\%) \\
\midrule
Kullback--Leibler & 76.87 \\
Chi-square & 76.06 \\
$\alpha$-divergence ($\alpha=1.5$) &  \bf 77.56 \\
Jensen--Shannon & 72.24\\
Squared Hellinger & 72.81\\
\bottomrule
\end{tabular}
\label{tab:acc_imagenet}
\end{table}

\subsection{Language modeling}
\label{subsec:lm}

We compare the performance of different $f$-divergence-based loss functions for training language models (LMs).  To provide a thorough evaluation, we consider three common LM training strategies: (i) pretraining, (ii) supervised fine-tuning (SFT), and (iii) distillation-based fine-tuning.  All three approaches optimize a next-token prediction loss:
\begin{equation}
\EE_{(\x,\y)} \left[ \sum_{t=1}^{|\y|} \ell_f(h_{\w}(\x, \y_{<t}), \phi(y_t); \q)\right], \label{eq:loss-sft-and-distillation}
\end{equation}
where $h_\w$ is the LM that outputs logits with parameters $\w$, $|\y|$ is the length of the text sequence $\y$, and the reference measure $\q = \ones$ is a unitary prior; cf.~Equation \eqref{eq:f-div-loss}.
Pretraining and SFT use one-hot vectors $\phi(y_t) \in \{\e_1, \ldots, \e_k\}$ as the targets; distillation uses soft labels $\phi(y_t) \in \triangle^k$ obtained as the next-token probability of a teacher model.


\paragraph{Pretraining with $f$-divergence generated losses.} We used the same pretraining method as the NanoDO models \citep{nanodo2024, wortsman2024smallscale}, a set of well-tuned decoder-only transformer models trained on the public C4 dataset \citep{raffel2020exploring}. 
Specifically, we follow the training set up of the 1.2B parameter model from \citet{wortsman2024smallscale}; we use $250$B tokens of C4 to train the model, so that the model is Chinchilla optimal \citep{Hoffmann2022training}. Our pretraining results (see Appendix~\ref{appendix:nanodo-pretraining}) echo our main findings from the ImageNet experiments in Section \ref{subsec:imagenet}. 
Specifically, losses based on $\alpha$-divergence, with $\alpha=1.5$, performed best for predicting the next word, slightly outperforming the standard KL divergence loss.

\paragraph{Learning $\q$.}

Since our loss functions take a reference distribution $\q$ and we can compute gradients with respect to $\q$, a natural idea is to try to learn it from data. As a negative result, using the NanoDO implementation above on the C4 dataset, we found that learning $\q$ does not seem to help improve performance compared to $\q=\ones$, both when using the KL divergence and when using the $\alpha=1.5$ divergence.

\paragraph{SFT and distillation with $f$-divergence generated losses.}


Instead of pretraining LMs from scratch, practitioners often fine-tune readily available, open-weight models pretrained with standard cross-entropy. This prompts a question: Can we effectively fine-tune these cross-entropy pretrained models using different $f$-divergence generated losses? To explore this, we evaluated $f$-divergence generated losses with two common finetuning methods, SFT and distillation, on a text summarization task \citep{narayan2018dont}.
For SFT, we use a pretrained T5-base model \citep{raffel2020exploring} with 250M parameters and train it on the XSum dataset \citep{narayan2018dont} with a next-token prediction loss \eqref{eq:loss-sft-and-distillation}. Traditionally, the cross-entropy loss is used as the next-token prediction loss, which is equivalent to KL-divergence. In our experiments, however, we evaluate more general $f$-divergence generated losses.


Distillation uses a large teacher model's class probabilities as soft labels to train a smaller student model \citep{hinton2015distilling}. In the LM context, such soft labels are the teacher model's next-token probabilities, which can be more effective than one-hot labels as they provide richer information about the likely next tokens. We use a T5-XL model (800M parameters) as the teacher and a T5-base model (250M parameters) as the student; prior to distillation, both models were SFT'ed on XSum as in \citet{agarwal2024policy}. We then fit the student model with the loss \eqref{eq:loss-sft-and-distillation}, where $\phi(y_t)$ in the loss are the soft labels produced by the teacher model.

\begin{figure}[t]
    \centering
    \includegraphics[width=0.45\textwidth]{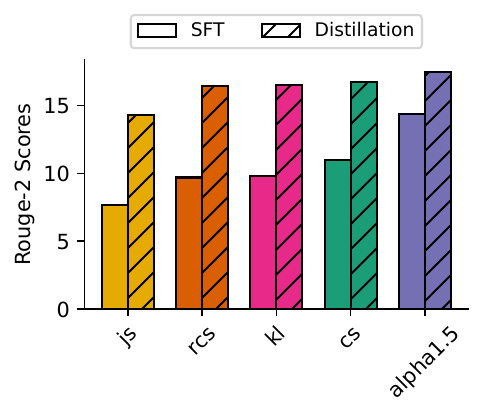}
    \vspace{-0.5cm}
    \caption{Comparing the effect of different $f$-divergence generated losses used for SFT and distillation training of LMs. The $f$-divergence generated losses are Jensen-Shannon (js), reverse Chi-square (rcs), Kullback--Leibler (kl), Chi-square (cs), and $\alpha$-divergence with $\alpha=1.5$ (alpha1.5) \label{fig:xsum_results}}
\end{figure}


For both SFT and distillation, we adopted the XSum settings from \citet[Appendix A.3]{agarwal2024policy}.  Following \citet{chowdhery2023palm} and \citet{agarwal2024policy}, we evaluated summarization quality using the ROUGE-2 score \citep{lin2004rouge} with temperature-based sampling (temperature=1). Figure~\ref{fig:xsum_results} shows results for SFT (unhatched) and distillation (hatched). Distillation consistently outperformed SFT.  Critically, Chi-squared and $\alpha$-divergence ($\alpha=1.5$) losses outperformed KL divergence for both training paradigms.
\paragraph{Comparing $f$-softargmax variants for decoding.} 

\begin{figure}[ht]
    \centering
    \includegraphics[width=0.45\textwidth]{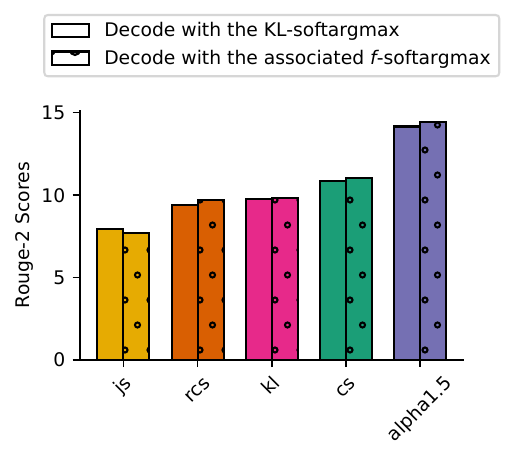}
    \vspace{-0.5cm}
    \caption{Decoding with the classical (KL-based) softargmax performs similarly to decoding with $f$-softargmax associated with the loss used for training. This suggests that the choice of $f$-divergence loss during training, not the decoding method, primarily drives performance differences. \label{fig:decode-with-different-softargmaxes}}
\end{figure}

In addition to the loss function used at the training of LMs, another key factor that influences the performance of LMs is how one decode responses from them. While the training of a LM uses a fixed $f$-divergence loss, when decoding, one can choose different $f$-softargmax variants to turn LM logits into next-token probabilities for sampling. In our prior experiments of SFT and distillation (Figure~\ref{fig:xsum_results}), at decoding time we used the $f$-softargmax corresponding to each $f$-divergence loss function used at training time. But this raises a question: Do the performance differences primarily come from the $f$-divergence loss used during training or from the usage of different $f$-softargmax at decoding? To study this, we perform SFT with different $f$-divergence generated losses as before, but then decode from the trained LM using either the associated $f$-softargmax or the  classical (KL-based) softargmax. Figure~\ref{fig:decode-with-different-softargmaxes} shows the results. Decoding with the two types of $f$-softargmax variants yields nearly identical results. Indeed, note that the two bars of KL divergence are slightly different only due to the effect of random sampling; the observed performance differences in the case of other divergences are comparable to those seen in the KL case. This suggests that the $f$-divergence used during training primarily contributes to the performance differences of models.

\subsection{Summary of empirical findings}

We summarize the main take-aways from our experiments.
\begin{itemize}[topsep=0pt,itemsep=3pt,parsep=3pt,leftmargin=15pt]

\item We found that the loss function generated by the $\alpha$-divergence with $\alpha=1.5$ works well across all tasks we tried. While equivalent to Tsallis negentropy ($\alpha=1.5$) with a uniform reference measure \citep{blondel2020learning,entmax}, its effectiveness across several language modeling tasks is novel. Its good performance could intuitively come from the fact that it is a middle ground between the logistic loss ($\alpha=1$) and the sparsemax loss ($\alpha=2$).
Figure \ref{fig:app_imagenet} confirms that accuracy is maximized around $\alpha=1.5$.

\item We successfully fine-tuned LMs with various $f$-divergence generated losses, despite the fact the pretraining was carried out using the cross-entropy loss. This enables direct application of our $f$-divergence generated losses to pretrained, open-weight LMs.

\item We show that, during text generation, standard softargmax yielded good performance even when the model is fine-tuned with $f$-divergence generated losses. This is surprising because \eqref{eq:identify_of_indiscernibles} tells us that we should in principle use the $f$-softargmax operator associated with the loss. This empirical finding opens up the possibility to use our losses without changing the inference code, which is convenient when working with open-weight LMs.

\item Loss functions generated by $f$-divergences not considered before (Jensen-Shannon, squared Hellinger) did not lead to better accuracy. The KL and $\alpha=1.5$ generated losses consistently achieved the best results across tasks.
    
\end{itemize}

\section{Related work}

\paragraph{Loss functions based on $f$-divergences.}

\citet{nguyen2009surrogate} studied surrogate losses for binary classification based on $f$-divergences.
However, this approach is not straightforward to generalize to the multiclass setting,
as it requires multi-distribution extensions of $f$-divergences \citep{garcia2012divergences,duchi2018multiclass}.
Recently, $f$-divergences have been used for distillation in combination with a classical softargmax 
\citep{agarwal2024policy}. In our notation, this defines a loss $\thetav \mapsto D_f(\y, \softargmax(\thetav))$.
However, this does not result in a convex loss in $\thetav$.
\citet{novello2024f} developed deep learning objectives based on the variational formulation of $f$-divergences.
\citet{sbert2024entropies, sharma2021geometric, aminian2024robust} also defined entropies from $f$-divergence to measure uncertainty, homogeneity or to build losses robust to label errors as also done by \citet{zhu2023label}.
$f$-divergences have also been used to design GANs~\citep{nowozin2016f} or more generally provide alternate learning frameworks~\citep{li2016renyi}.

\citet{bregman_fdiv} briefly studied the idea of generating Bregman divergences from $f$-divergences. However, Bregman divergences need to be explicitly composed with a softargmax and this composition is again usually not convex.
In contrast, in our approach, each loss $\thetav \mapsto \ell_f(\thetav, \y; \q)$ is convex and is implicitly associated with the corresponding $\softargmax_f$ operator.


\paragraph{LM alignment with $f$-divergences.}

In RLHF, given a learned reward model $r \colon \cX \times \cY \to \RR$,
the optimal policy is
\begin{equation}
\p^\star(\cdot|\x) \coloneqq 
\argmax_{\p \in \triangle^{|\cY|}} \EE_{\y \sim \p}[r(\x, \y)] 
- \mathrm{KL}(\p, \q(\cdot|\x)),
\end{equation}
where 
$\x \in \cX$ is a given prompt
and
$\q(\cdot|\x) \in \triangle^{|\cY|}$ is a reference conditional distribution, 
usually a pre-trained model.
This approach can be generalized to $f$-divergences \citep{go2023aligning}.
In our notation, this can be written as
\begin{equation}
\p^\star(\cdot|\x) = \softargmax_f(r(\x, \cdot), \q(\cdot|\x)).
\end{equation}
To avoid learning a separate reward model altogether, we can use 
direct preference optimization (DPO) \citep{rafailov2024direct}, which was generalized to $f$-divergences \citep{wang2023beyond}. Proposition \ref{prop:computation} can be seen as generalization of \citep[Theorem 1]{wang2023beyond} that supports $0 \in \dom(f')$. 
In this paper, our focus was on the SFT and distillation steps, which usually precede RLHF.
Unlike these works, we propose to use the $f$-softargmax operator and the corresponding Fenchel--Young loss for pretraining, SFT and distillation in the space of tokens, making SGD-based optimization easy.

\section{Conclusion}

We proposed to use $f$-divergences as regularization for generating loss functions and associated operators ($f$-softmax, $f$-softargmax, $f$-softplus and $f$-sigmoid). Our proposal establishes a link between the logistic, sparsemax and entmax loss functions and the KL, Chi-square and $\alpha$ divergences, respectively. Thanks to this perspective, our proposal generalizes these losses to non-uniform class weights. Other choices of $f$-divergences allowed us to create entirely new loss functions.
Overall, we found that the loss function generated by the $\alpha=1.5$ divergence worked comparably or better to the cross-entropy loss on all tasks we tried.



\section*{Impact statement}

This paper explores $f$-divergence generated loss functions for classification or language modelling. We do not foresee any specific ethical or societal implications arising directly from this work.

\section*{Acknowledgments}
We thank the anonymous reviewers for their comments that helped improve the paper.





\bibliography{references}
\bibliographystyle{icml2025}

\newpage
\appendix
\onecolumn

\section{Experimental details and additional experiments}

\subsection{ImageNet classification \label{appendix:imagenet-classification}}

When using the KL divergence, our training yields 76.87\% accuracy on ImageNet. 
This result matches previous benchmarks of ResNet50 trained on ImageNet, such as \url{https://github.com/google/flax/tree/main/examples/imagenet}.

\begin{figure}[H]
    \centering
    \includegraphics[scale=0.5]{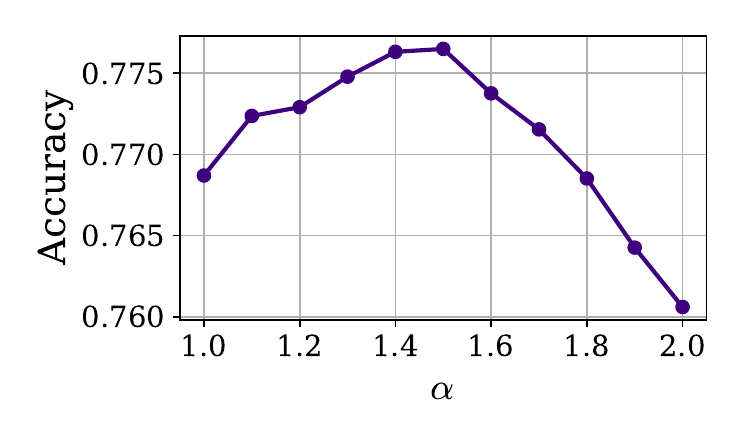}
    \caption{Validation accuracy on ImageNet when using the $\alpha$-divergence generated loss, with $\alpha \in [1., 2.]$. Accuracy is maximized near $\alpha = 1.5$.}
    \label{fig:app_imagenet}
\end{figure}

\subsection{Pretraining NanoDO models \label{appendix:nanodo-pretraining}}

The 1.2B NanoDO model's architecture and hyperparameters follow from the setup in \citet{wortsman2024smallscale}.  The model is similar to GPT2 \citep{radford2019language} but incorporates modern features like rotary positional embeddings \citep{Su2024roformer} and qk-layernorm \citep{Dehghani2023scaling}.
To evaluate the trained model, we use the next-token prediction accuracy on the validation dataset. Our results (Table~\ref{tab:acc_nanodo}) show that $\alpha$-divergence is on par, and slightly more performant, than the classifical KL approach.

\begin{table}[ht]
\centering
\caption{Next-token prediction accuracy}
\begin{tabular}{lc}
\toprule
Divergence & Accuracy (\%) \\
\midrule
Kullback--Leibler & 48.66 \\
Chi-square & 46.75 \\
$\alpha$-divergence ($\alpha=1.5$) &  \bf{48.70} \\
\bottomrule
\end{tabular}
\label{tab:acc_nanodo}
\end{table}

\subsection{Standard deviation of accuracy results over random seeds}

\begin{table}[H]
\centering
\caption{Mean, standard deviation, min, max accuracies over 5 random seeds for the ImageNet classification results.}
\label{tab:imagenet_further_updated}
\begin{tabular}{lcccc}
\toprule
Divergence & Mean & Standard Deviation & Min & Max \\
\midrule
Kullback--Leibler & 0.7684 & 0.0007 & 0.7676 & 0.7692 \\
Chi--square & 0.7604 & 0.0015 & 0.7587 & 0.7621 \\
$\alpha$-divergence ($\alpha=1.5$) & {\bf 0.7758} & 0.0013 & 0.7743 & 0.7776 \\
Jensen--Shannon & 0.7246 & 0.0014 & 0.723 & 0.7266 \\
Squared Hellinger & 0.7281 & 0.0008 & 0.7272 & 0.729 \\
\bottomrule
\end{tabular}
\end{table}

\begin{table}[H]
\centering
\caption{Mean, standard deviation, min, max Rouge-2 scores over 5 random seeds for the supervised fine-tuning (SFT) results.}
\label{tab:sft_updated}
\begin{tabular}{lcccc}
\toprule
Divergence & Mean & Standard Deviation & Min & Max \\
\midrule
Kullback--Leibler & 9.77 & 0.02 & 9.75 & 9.8 \\
Chi--square & 11.15 & 0.1 & 11.02 & 11.31 \\
$\alpha$-divergence ($\alpha=1.5$) & {\bf 14.27} & 0.04 & 14.2 & 14.32 \\
Jensen--Shannon & 7.95 & 0.08 & 7.87 & 8.07 \\
Reverse Chi--squared & 9.55 & 0.1 & 9.44 & 9.7 \\
\bottomrule
\end{tabular}
\end{table}

\begin{table}[H]
\centering
\caption{Mean, standard deviation, min, max Rouge-2 scores over 5 random seeds for the distillation results.}
\label{tab:distillation_updated}
\begin{tabular}{lcccc}
\toprule
Divergence & Mean & Standard Deviation & Min & Max \\
\midrule
Kullback--Leibler & 16.64 & 0.05 & 16.57 & 16.71 \\
Chi--square & 14.17 & 0.09 & 14.01 & 14.26 \\
$\alpha$-divergence ($\alpha=1.5$) & {\bf 17.43} & 0.13 & 17.19 & 17.6 \\
Jensen--Shannon & 16.51 & 0.06 & 16.43 & 16.6 \\
Reverse Chi--squared & 16.3 & 0.05 & 16.25 & 16.38 \\
\bottomrule
\end{tabular}
\end{table}

\subsection{Computational cost}\label{app:comp_cost}

\begin{figure}[H]
    \centering
    \includegraphics[scale=0.6]{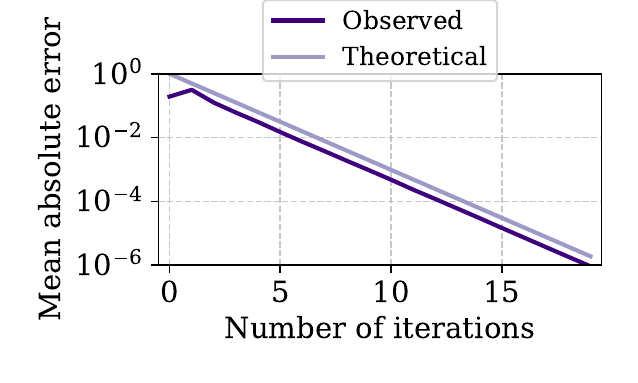}
    \caption{Error for computing the root in Proposition \ref{prop:computation} using our bisection based Algorithm \ref{algo:bisect} as a function of the number of iterations. We use the output with $30$ iterations as a proxy for the true root. We compare the measured error (dark purple) with the theoretical $2^{-t}$ error (light purple). This experiment is run on TPU.}
    \label{fig:bisection}
\end{figure}

\paragraph{Bisection convergence.}
We experimentally validate the exponential convergence of our proposed Algorithm \ref{algo:bisect}. Our results are in Figure \ref{fig:bisection} and confirm the theoretical convergence rate.

\begin{figure}[H]
    \centering
    \includegraphics[scale=0.6]{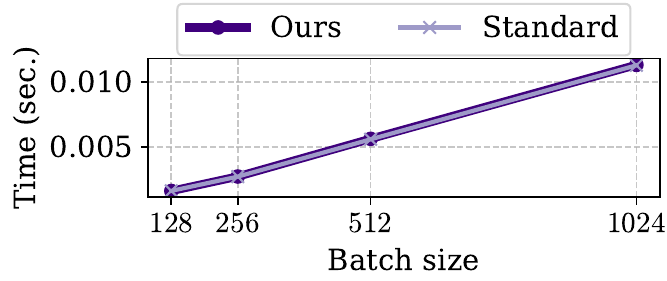}
    \caption{Profiled time for an input of shape $(b, 224, 224, 3)$ to be processed by a ResNet18 followed by $f$-softargmax corresponding the $1.5$-divergence (dark purple) or the standard Kullback–Leibler divergence (light purple, in which case we use the standard JAX implementation) for different batches size $b$. This experiment is run on TPU.}
    \label{fig:resnet}
\end{figure}

\paragraph{Overall cost.} We propose an experiment validating that our proposed losses generated by $f$-divergences and associated operators do not affect the runtime of standard learning pipelines.
For this, we consider a Residual Network \citep{he2016deep} ResNet18, with either the standard softargmax or the softargmax associated with the $\alpha =1.5$ divergence. For different batch sizes, we profile the time needed for running both the model and the $f$-softargmax. Results in Figure \ref{fig:resnet} show that the runtimes are approximately equal.

\subsection{Impact of prior distribution on sigmoids}

\begin{figure}[H]
    \centering
    \includegraphics[scale=0.5]{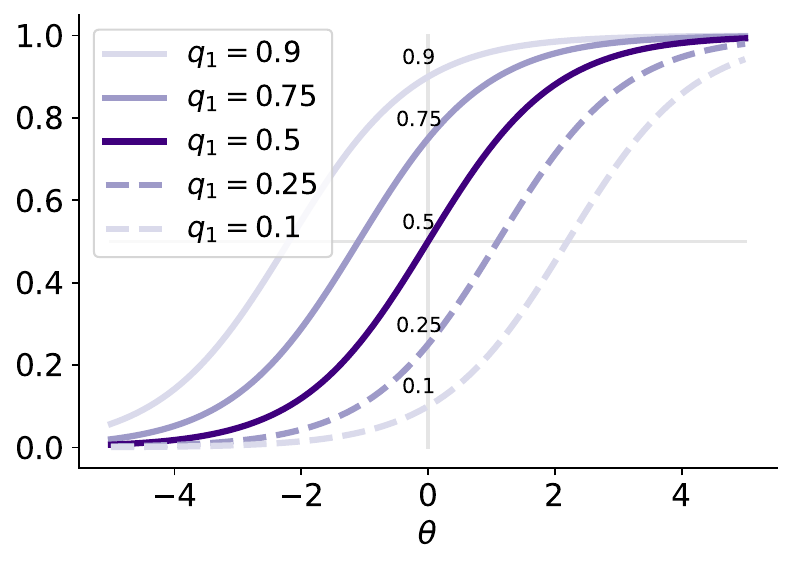}
    \caption{Illustration of $\theta \mapsto \sigmoid_f(\theta; (1-q_1, q_1))$ for $q_1 \in \{0.1, 0.25, 0.5, 0.75, 0.9\}$ and for $f(u) = u \log u$, the generating function of the KL divergence. We see that
    $\sigmoid_f(\theta; (1-q_1, q_1))$ at $\theta=0$ is equal to $q_1$. Intuitively, if a model is uncertain and produces a value of $\theta=0$, then the model outputs $q_1$, the prior probability of the positive class.}
    \label{fig:sigmoid}
\end{figure}

\begin{figure}[p]
    \centering
    \includegraphics[width=0.65\linewidth]{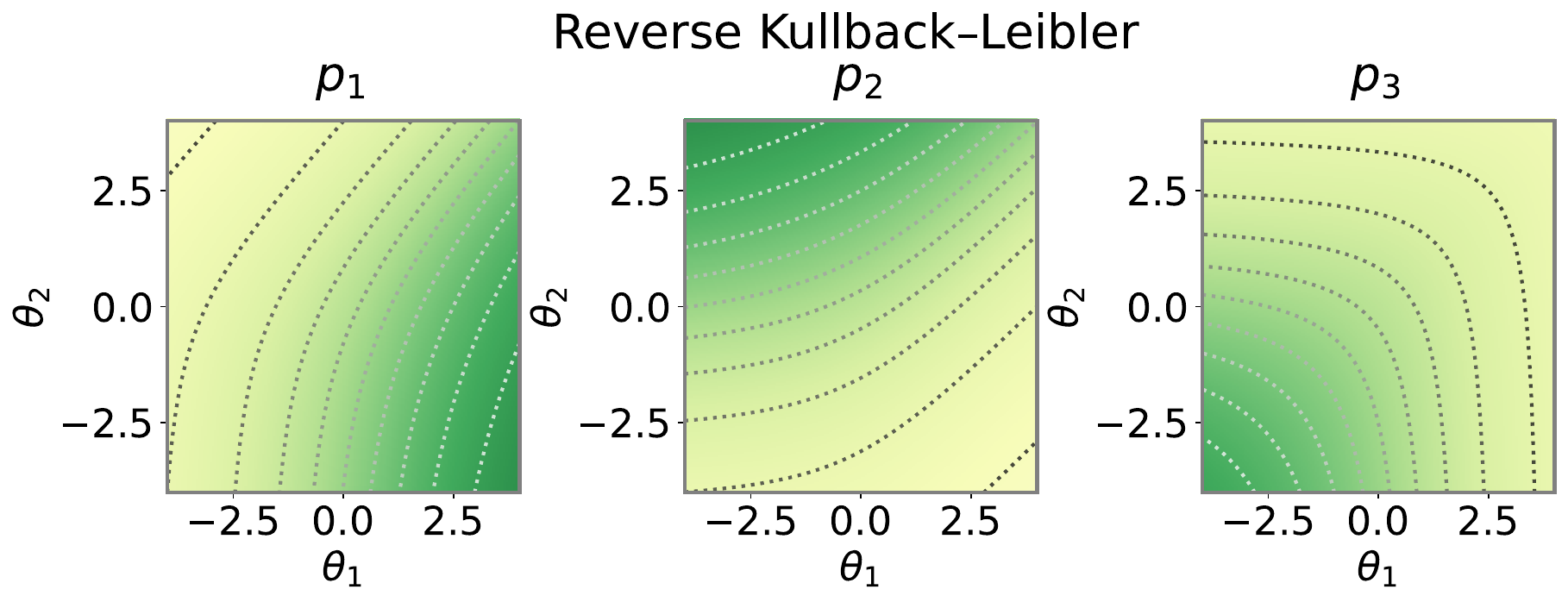}
    \includegraphics[width=0.65\linewidth]{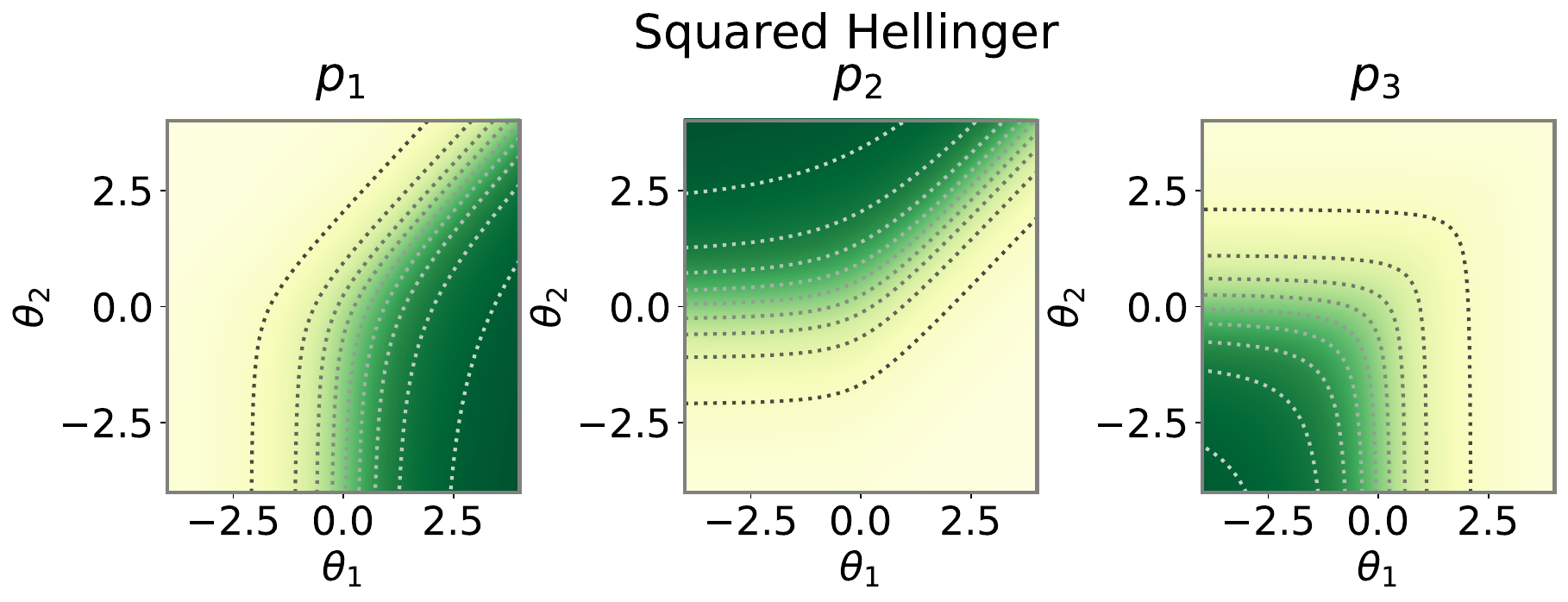}
    \includegraphics[width=0.65\linewidth]{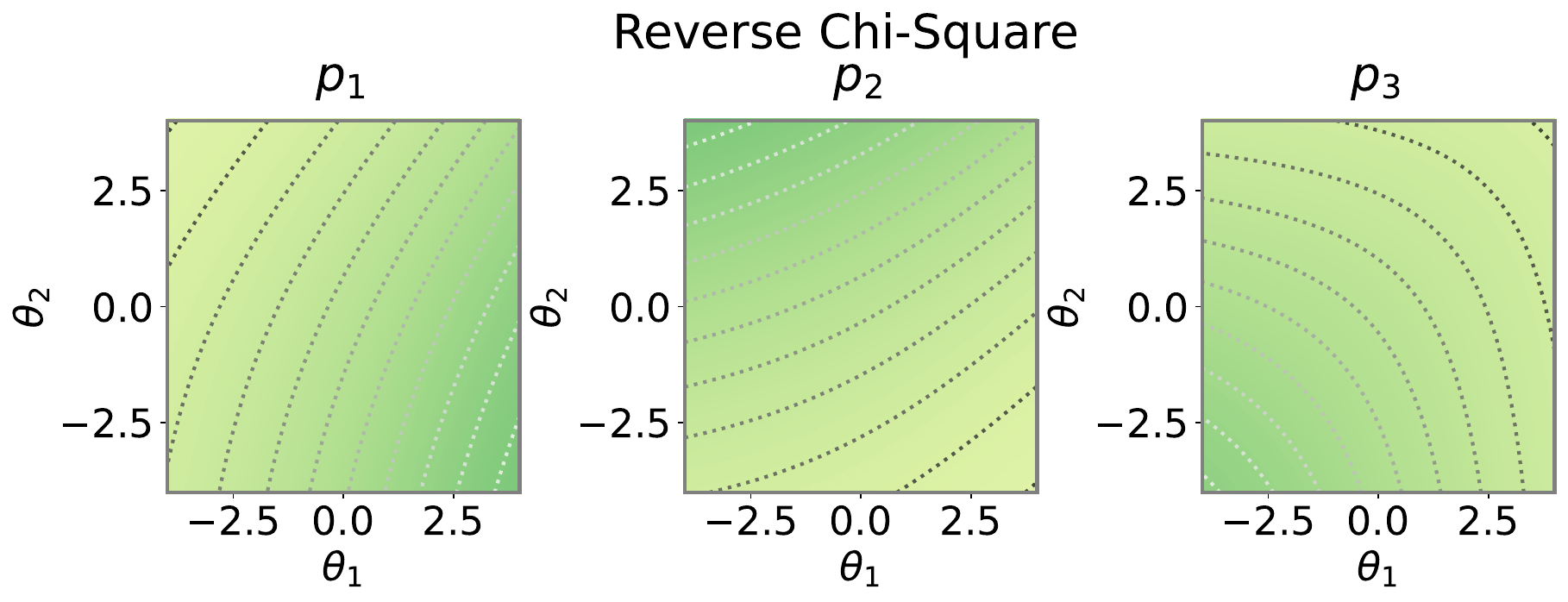}
    \includegraphics[width=0.65\linewidth]{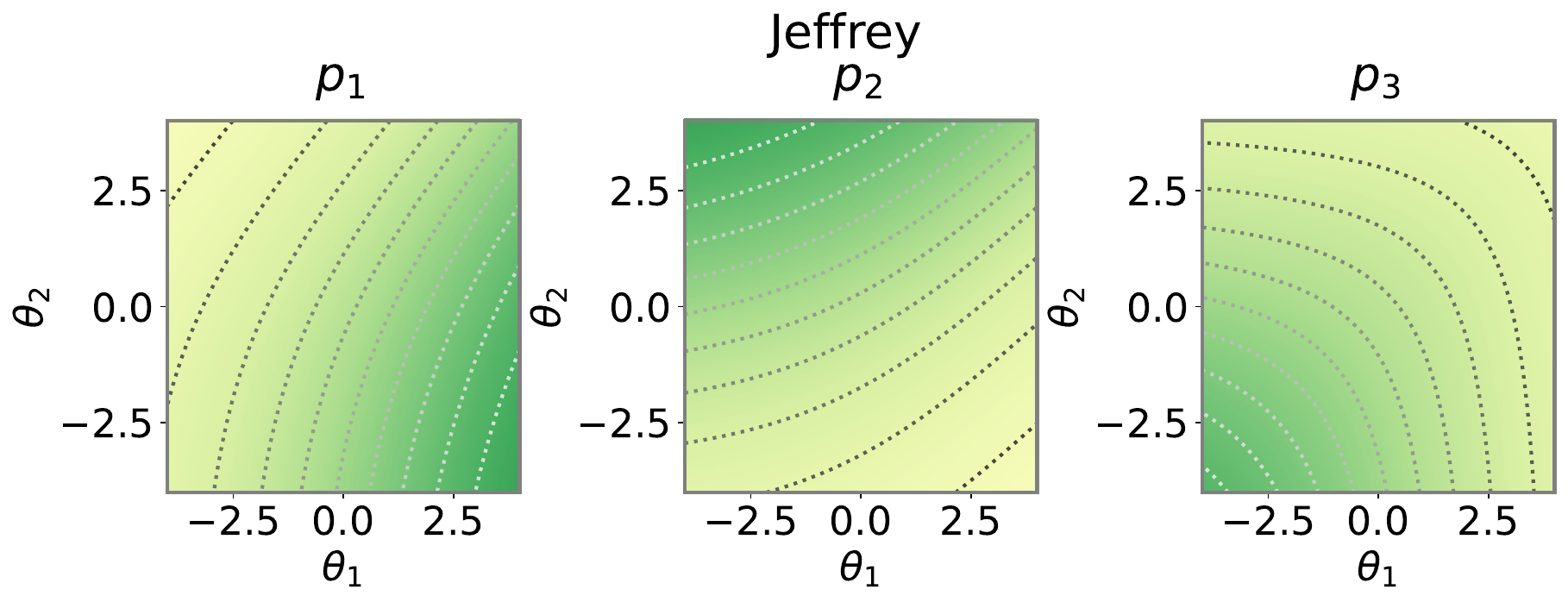}
    \includegraphics[width=0.65\linewidth]{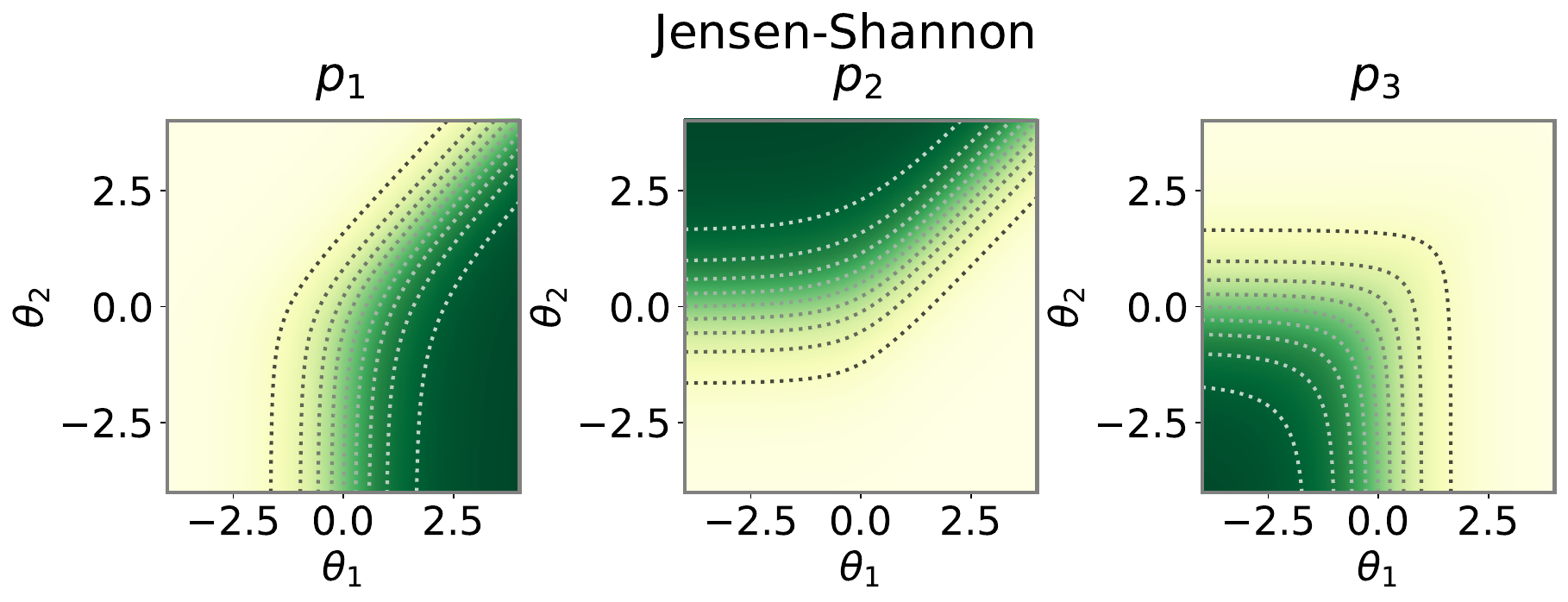}
    \hspace*{30pt}\includegraphics[width=0.3\linewidth]{figures/softargmax_colorbar.pdf}
    \vspace{-0.4cm}
    \caption{Illustration of $(p_1, p_2, p_3) = \softargmax_f(\theta_1, \theta_2, 0; \q)$ 
    when varying $\theta_1, \theta_2 \in \RR$ for diverse $f$-divergences 
    and with $\q = (1, 1, 1)$.}
    \label{fig:softargmax2}
\end{figure}

\begin{figure}[p]
    \centering
    \includegraphics[width=0.7\linewidth]{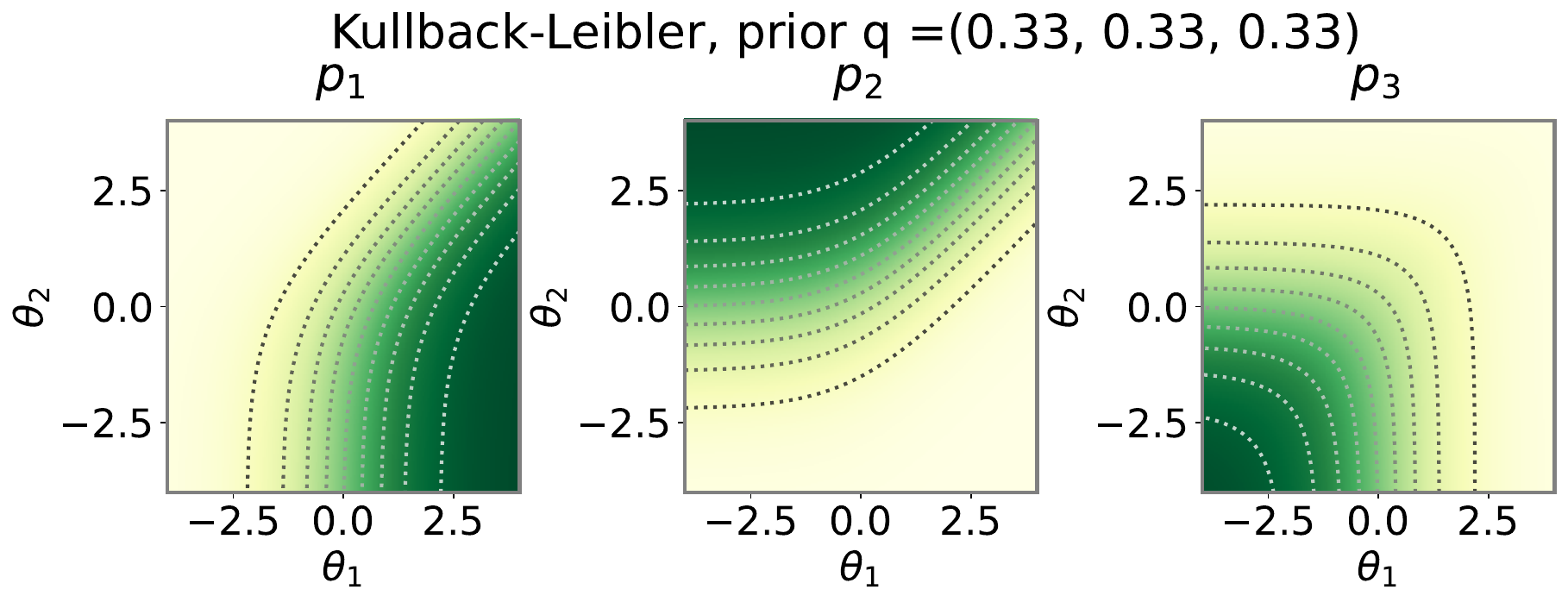}
    \includegraphics[width=0.7\linewidth]{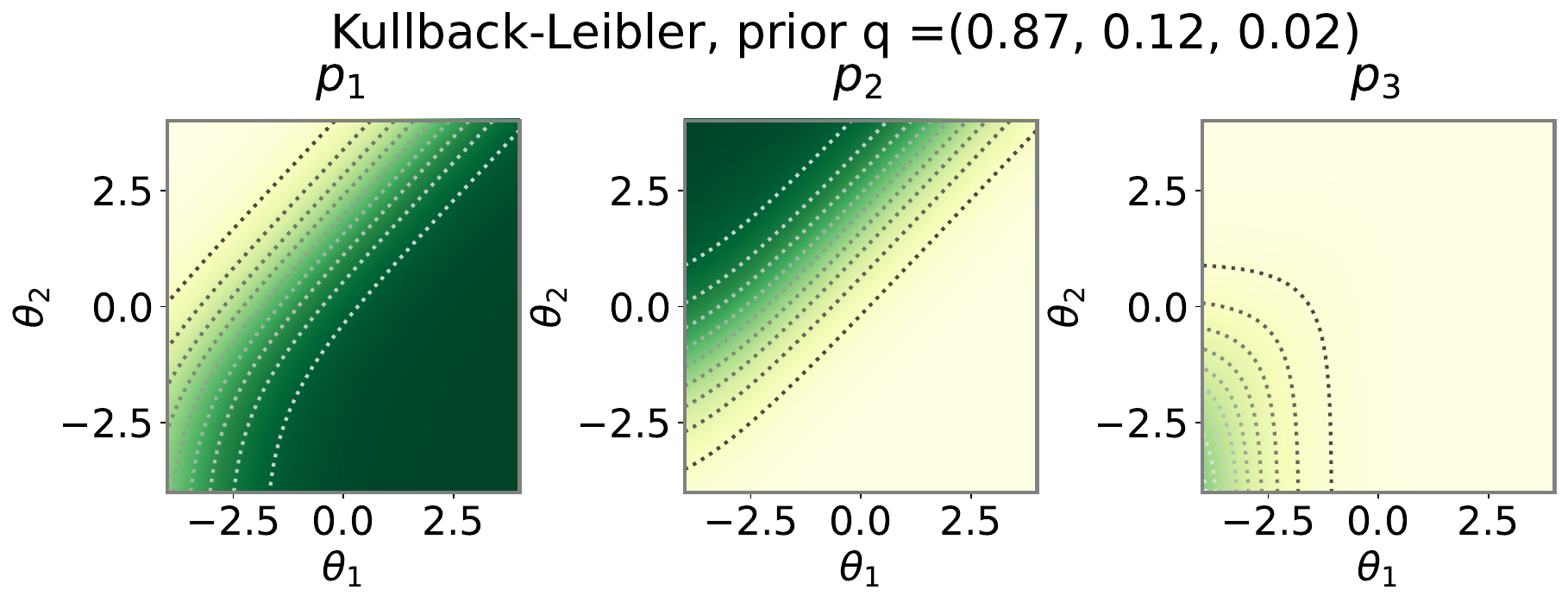}
    \includegraphics[width=0.7\linewidth]{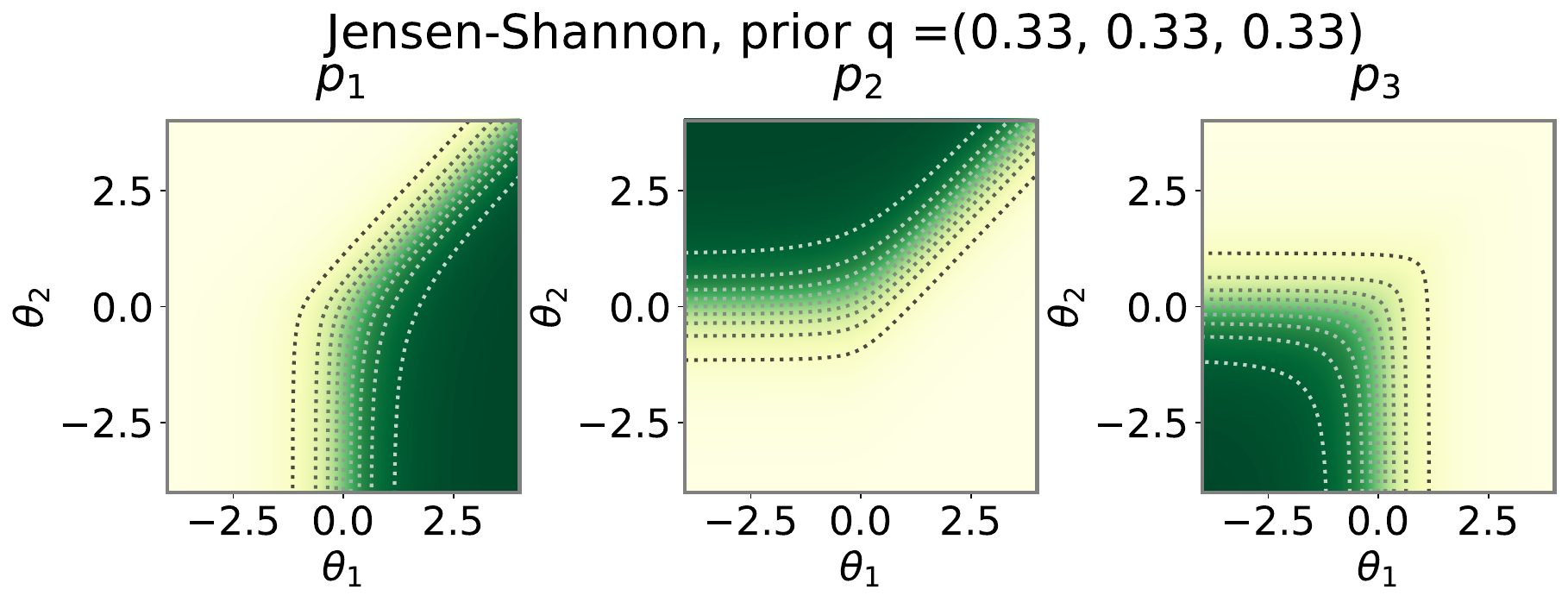}
    \includegraphics[width=0.7\linewidth]{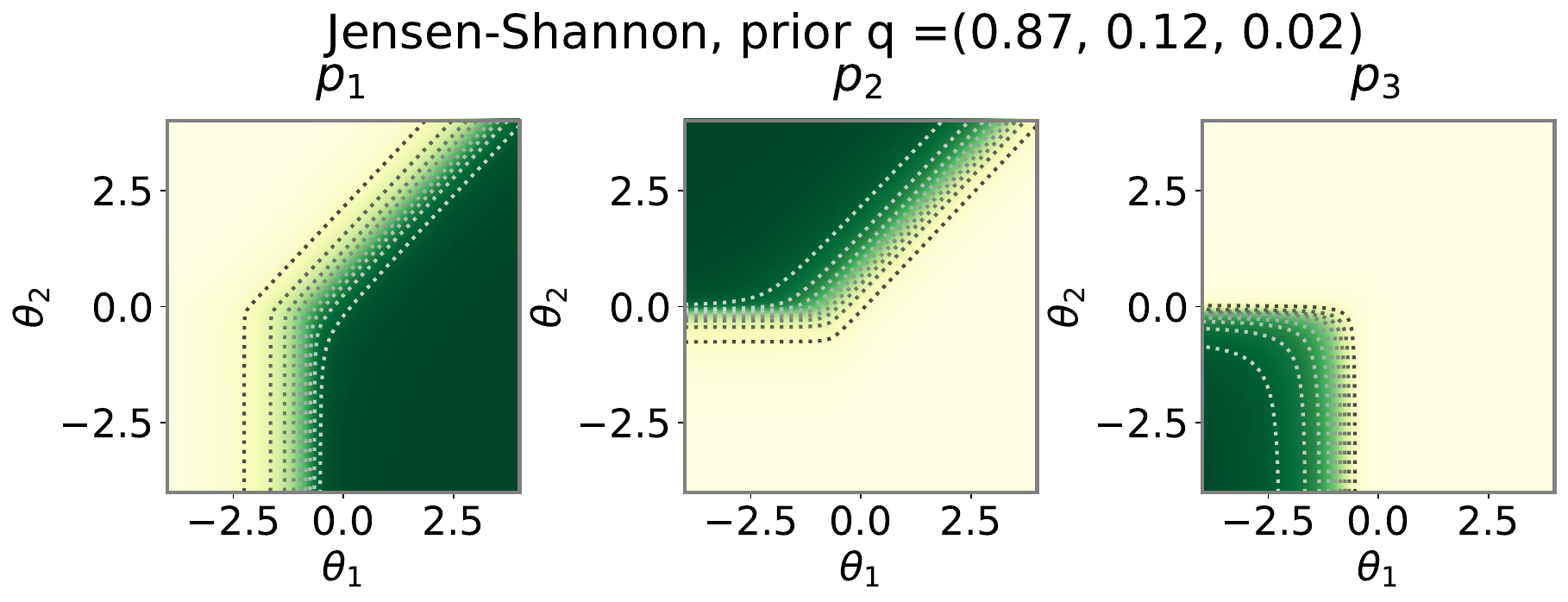}
    \hspace*{30pt}\includegraphics[width=0.3\linewidth]{figures/softargmax_colorbar.pdf}
    \caption{Illustration of $(p_1, p_2, p_3) = \softargmax_{f}(\theta_1, \theta_2, 0; \q)$ 
    when varying $\theta_1, \theta_2 \in \RR$ for diverse $f$-divergences and
    priors $\q$.}
    \label{fig:softargmax_non_uniform}
\end{figure}

\clearpage

\section{Examples of $f$-divergences \label{app:f_div_examples}}

\begin{itemize}
    \item KL
    \begin{itemize}
        \item[$\bullet$] $f(u) = u \log u$, $\dom(f) = \RR_+$
        \item[$\bullet$] $f'(u) = \log u + 1$
        \item[$\bullet$] $f_*(v) = \exp(v - 1)$, $\dom(f_*) = \RR$
        \item[$\bullet$] $f_*'(v) = \exp(v - 1)$
        \item[$\bullet$] $D_f(p, q) = \mathrm{KL}(p,q) \coloneqq \langle p, \log p \rangle - \langle p, \log q \rangle$
    \end{itemize}
    
    \item Generalized KL
    \begin{itemize}
        \item[$\bullet$] $f(u) = u \log u - (u - 1)$, $\dom(f) = \RR_+$
        \item[$\bullet$] $f'(u) = \log u$
        \item[$\bullet$] $f_*(v) = \exp(v) - 1$, $\dom(f_*) = \RR$
        \item[$\bullet$] $f_*'(v) = \exp(v)$
        \item[$\bullet$] $D_f(p, q) = \mathrm{GKL}(p,q) 
        \coloneqq \langle p, \log p \rangle - \langle p, \log q \rangle - \langle p, 1 \rangle + \langle q, 1 \rangle$
    \end{itemize}
    
    \item Reverse KL
    \begin{itemize}
         \item[$\bullet$] $f(u) = -\log u$, $\dom(f) = \RR_+$
         \item[$\bullet$] $f'(u) = -1/u$
         \item[$\bullet$] $f_*(v) = -1 - \log(-v)$, $\dom(f_*) = \RR_-$ (\cf Proposition \ref{prop:conjugate_reverse_kl})
         \item[$\bullet$] $f_*'(v) = -1/v$
         \item[$\bullet$] $D_f(p, q) = \mathrm{KL}(q,p)$
     \end{itemize}
     
    
    \item Jeffreys 
    \begin{itemize}
         \item[$\bullet$] $f(u) = (u-1) \log u$, $\dom(f) = \RR_+$
         \item[$\bullet$] $f'(u) = \log u + 1 - \frac{1}{u}$
         \item[$\bullet$] $f_*(v) = \frac{1}{W(\exp(1 - v))} + \log \frac{1}{W(\exp(1 - v))} - 1$, $\dom(f_*) = \RR$
         (\cf Proposition \ref{prop:conjugate_jeffrey})
         \item[$\bullet$] $f_*'(v) = \frac{1}{W(\exp(1 - v))}$ 
         \item[$\bullet$] $D_f(p, q) = \mathrm{KL}(p,q) + \mathrm{KL}(q,p) = \langle \log p - \log q, p - q \rangle$
         \item[$\bullet$] Remark: $W$ is the Lambert function
     \end{itemize}
     
    \item Jensen-Shannon
    
    \begin{itemize}
         \item[$\bullet$] $f(u) = u \log u - (u+1) \log\left(\frac{u+1}{2}\right)$, $\dom(f) = \RR_+$
         \item[$\bullet$] $f'(u) = \log u - \log\left(\frac{u+1}{2}\right) = \log\left(\frac{2u}{u+1}\right)$
         \item[$\bullet$] $f_*(v) = -\log(2 - \exp(v))$, $\dom(f_*) =(-\infty, \log 2)$
         (\cf Proposition \ref{prop:conjugate_jensen_shannon})
         \item[$\bullet$] $f'_*(v) = \frac{\exp(v)}{2 - \exp(v)} = \frac{1}{2 \exp(-v) - 1}$
         \item[$\bullet$] $D_f(p, q) 
         = 2 \cdot \mathrm{JS}(p,q) 
         \coloneqq  \mathrm{KL}(p, m) + \mathrm{KL}(q, m)
         = \left\langle p, \log\left(\frac{2p}{p+q}\right)\right\rangle + \left\langle q, \log\left(\frac{2q}{p+q}\right)\right\rangle$,
         where $m \coloneqq \frac{1}{2}(p + q)$. 
     \end{itemize}

    \item Squared Hellinger 
    \begin{itemize}
         \item[$\bullet$] $f(u) = (\sqrt{u} - 1)^2 = u - 2 \sqrt{u} + 1$, $\dom(f) = \RR_+$
         \item[$\bullet$] $f'(u) = 1 - \frac{1}{\sqrt{u}}$
         \item[$\bullet$] $f_*(v) = \frac{v}{1-v}$, $\dom(f_*) = (-\infty, 1)$ 
         (\cf Proposition \ref{prop:conjugate_squared_hellinger})
         \item[$\bullet$] $f_*'(v) = \frac{1}{(1-v)^2}$
         \item[$\bullet$] $D_f(p, q) = 2 \cdot \mathrm{SH}(p,q) = \sum_{j=1}^k (\sqrt{p_j} - \sqrt{q_j})^2$
     \end{itemize}
     
    \item Chi-squared divergence (\aka Pearson divergence)
    \begin{itemize}

         \item[$\bullet$] $f(u) = \frac{1}{2}(u^2 - 1)$, $\dom(f) = \RR$
         \item[$\bullet$] $f'(u) = u$
         \item[$\bullet$] $f_*(v) =  \frac{1}{2}(v^2 + 1)$, $\dom(f_*) = \RR$
         (\cf Proposition \ref{prop:conjugate_chi_square})
         \item[$\bullet$] $f_*'(v) = v$
         \item[$\bullet$] $D_f(p, q) = \frac{1}{2} \chi^2(p, q) \coloneqq \frac{1}{2} \sum_{j=1}^k \frac{(p_j - q_j)^2}{q_j}$
         
         \item[$\bullet$] Remark: other possible generating functions are $f(u) = \frac{1}{2} (u-1)^2$ or $f(u) = \frac{1}{2}(u^2 - u)$. We choose $f(u) = \frac{1}{2}(u^2 - 1)$ as it satisfies $f'(0) = 0$.
     \end{itemize}
     
    \item Reverse Chi-squared divergence (\aka Neyman divergence)
    \begin{itemize}
         \item[$\bullet$] $f(u) = \frac{1}{2}(\frac{1}{u} - 1)$, $\dom(f) = \RR_+$
         \item[$\bullet$] $f'(u) = -\frac{1}{2u^2}$
         \item[$\bullet$] $f_*(v) = -\sqrt{-2v} + \frac{1}{2}$, $\dom(f_*) = \RR_-$
         (\cf Proposition \ref{prop:conjugate_reverse_chi_square})
         \item[$\bullet$] $f_*'(v) = \frac{1}{\sqrt{-2v}}$
         \item[$\bullet$] $D_f(p, q) = \frac{1}{2} \chi^2(q, p)$
     \end{itemize}
     
     \item $\alpha$-divergences
    \begin{itemize}
         \item[$\bullet$] 
         $f(u) = \begin{cases} 
         \frac{(u^\alpha - 1) - \alpha(u - 1)}{\alpha(\alpha-1)} + \iota_{\RR_+}(u) &\mbox{ if } \alpha \neq 1 \\ 
         u \log u - (u - 1) &\mbox{ if } \alpha = 1 
         \end{cases}$, 
         $\dom(f) = \RR_+$
         
         \item[$\bullet$] $f'(u) = \log_\alpha(u) \coloneqq \begin{cases}
\frac{u^{\alpha-1} - 1}{\alpha - 1} &\mbox{ if } \alpha \neq 1 \\
\log u &\mbox{ if } \alpha = 1
\end{cases}$

         \item[$\bullet$] $f_*(v) = \begin{cases}
\frac{(1+(\alpha-1)v)^{\frac{\alpha}{\alpha-1}} - 1}{\alpha} &\mbox{ if } \alpha \in (0, 1) \\
\exp(v) &\mbox{ if } \alpha = 1 \\
\frac{(1+(\alpha-1)v)_+^{\frac{\alpha}{\alpha-1}} - 1}{\alpha} & \mbox{ if } \alpha \in (1, +\infty), \ \mbox{where} \ x_+ = \max(x, 0)
\end{cases}$, (\cf Proposition \ref{prop:conjugate_alpha_div})

        \item[$\bullet$] $\dom f_* = \begin{cases}
        (-\infty, -\frac{1}{\alpha-1}] & \mbox{ if } \alpha \in (0, 1) \\
        \RR & \mbox{ if } \alpha \geq 1
        \end{cases}$

         \item[$\bullet$] $f_*'(v) = \exp_\alpha(v) \coloneqq \begin{cases}
(1 + (\alpha - 1)v)^{\frac{1}{\alpha-1}} &\mbox{ if } \alpha \in (0, 1) \\
\exp v &\mbox{ if } \alpha = 1 \\
(1 + (\alpha - 1)v)_+^{\frac{1}{\alpha-1}} &\mbox{ if } \alpha \in (1, +\infty)
\end{cases}$
         
         \item[$\bullet$] $D_f(p, q)
= \frac{1}{\alpha} (\langle p, \log_\alpha(p/q) \rangle - \langle p, 1 \rangle + \langle q, 1 \rangle)
= \frac{1}{\alpha(\alpha-1)} \left(\langle p^\alpha, q^{\alpha-1} \rangle - \alpha \langle p, 1 \rangle + (\alpha-1) \langle q, 1 \rangle\right)$

\item[$\bullet$] Remark: By analogy with the KL divergence case, the $\alpha$-divergence can be derived by setting $f'(u) \coloneqq \log_\alpha(u)$ and by integrating. The constant of integration is chosen so as to satisfy $f(1) = 0$.

     \end{itemize}  
     
     
\end{itemize}

All functions $f$ above can easily be shown to be strictly convex and differentiable.
Note that both properties (strict convexity and differentiability) are defined on the interior of the domain of definition of the function, and therefore non-differentiability on its border (typically $0$ on the left) does not matter to assess these properties.

\section{Proofs}

\subsection{Conjugates}

In this section, we derive the conjugate of $f$,
\begin{equation}
f_*(v) \coloneqq \max_u uv - f(u)
\end{equation}
for the divergences listed in Appendix \ref{app:f_div_examples}.

\paragraph{Reverse KL divergence.}

\begin{proposition}
\label{prop:conjugate_reverse_kl}
The conjugate of $f(u)=-\log(u)$ for $u \in \dom f = (0, +\infty)$ is $f_*(v) = -1 - \log(-v)$ for $v\in \dom f_*= (-\infty, 0) $.
\end{proposition}
\begin{proof}
For $v > 0$, the function $u\mapsto uv - f(u)$ is unbounded above by taking $u \rightarrow + \infty$.
For $v<0$, we have $\lim_{u\rightarrow 0} \{uv - f(u)\} = \lim_{u \rightarrow +\infty} \{uv -f(u)\} = -\infty $. Hence a maximizer $u^\star$ of $u\mapsto uv - f(u)$ exists and is characterized by $v = f'(u^\star)$. Since $f'(u) = -1/u$, we have
\begin{align*}
    v = f'(u) \iff v = -1/u \iff u = -1/v = f_*'(v).
\end{align*}
We then get
\begin{align*}
    f_*(v) & = - 1 + \log(-1/v) = -1 - \log(-v).
\end{align*}

\end{proof}

\paragraph{Jeffreys divergence.}

\begin{proposition}
\label{prop:conjugate_jeffrey}
The conjugate of $f(u) = (u-1) \log u$ for $u \in \dom f = (0, +\infty)$  is 
$f_*(v) = \frac{1}{W(\exp(1 - v))} + \log \frac{1}{W(\exp(1 - v))} - 1$ for $v \in \dom f_* = \RR$, where $W$ is the Lambert W function.
\end{proposition}
\begin{proof}
For any $v\in \RR$, we have $\lim_{u \rightarrow 0} \{uv - f(u)\} = \lim_{u \rightarrow +\infty} \{uv - f(u)\} = -\infty$. Therefore, for any $v \in \RR$, the function $u \mapsto uv - f(u)$ has a maximizer $u^\star$ characterized by $v = f'(u^\star)$.

Using $f'(u) = \log u + 1 - \frac{1}{u}$, we get
\begin{equation}
v = f'(u) = \log u + 1 - \frac{1}{u}
\iff 1 - v = 1/u + \log(1/u)
\iff \exp(1 - v) = \exp(1/u) 1/u.
\end{equation}
We recall the definition of the Lambert $W$ function.

For $z \ge 0$, $W(z)$ is the inverse of the function $g(w) = w \exp(w)$, i.e., $W(z) = g^{-1}(z)$ and $W^{-1}(w) = g(w)$. We then obtain
\begin{equation}
u = \frac{1}{W(\exp(1 - v))} = f_*'(v).
\end{equation}
Furthermore, we have 
\begin{equation}
uv = u \log u + u - 1
\end{equation}
and therefore, we get
\begin{align}
f_*(v) 
&= uv - u \log u + \log u \\
&= u + \log u - 1 \\
&= \frac{1}{W(\exp(1 - v))} + \log \frac{1}{W(\exp(1 - v))} - 1.
\end{align}
\end{proof}

\paragraph{Jensen-Shannon divergence.}

\begin{proposition}
\label{prop:conjugate_jensen_shannon}
The conjugate of
$f(u) = u \log u - (u+1) \log\left(\frac{u+1}{2}\right)$
for $u\in \dom f = (0, +\infty)$
is
$f_*(v) = \log u - v = -\log(2 - \exp(v))$,
for $v \in \dom f_* = (-\infty, \log 2)$.
\end{proposition}
\begin{proof}
We have for $u\rightarrow +\infty$, 
$uv - u \log u + (u+1) \log((u+1)/2) \sim uv - u\log u + u\log u/2 \sim uv - u \log 2$,
where $\sim$ denotes asymptotic equivalence.
Hence for $v > \log 2$, we have $\lim_{u \rightarrow + \infty} uv - f(u) = +\infty$. For $v < \log 2$, we have $\lim_{u \rightarrow + \infty} \{uv - f(u)\} = \lim_{u \rightarrow 0} \{uv - f(u)\}= -\infty$. Hence, for $v < \log 2$, a maximizer $u^\star$ of $u\mapsto uv - f(u)$ exists and is characterized by $v = f'(u^\star)$.

Using $f'(u) = \log\left(\frac{2u}{u+1}\right)$, we get
\begin{equation}
v = f'(u)
\iff \exp(v) = \frac{2u}{u+1}
\iff u = \frac{\exp(v)}{2 - \exp(v)} 
= \frac{\exp(0)}{2 \exp(-v) - \exp(0)} = f'_*(v).
\end{equation}
Let us now compute $f_*(v) = uv - u \log u + (u+1) \log\left(\frac{u+1}{2}\right)$.
We observe that
\begin{equation}
u+1 = \frac{2}{2 - \exp(v)}
\iff
\frac{u+1}{2} = \frac{u}{\exp(v)}
\iff
\log\left(\frac{u+1}{2}\right)
= \log u - v.
\end{equation}
Therefore
\begin{equation}
(u+1)\log\left(\frac{u+1}{2}\right)
= u (\log u - v) + \log u - v
= u \log u - uv + \log u - v.
\end{equation}
Therefore
\begin{equation}
f_*(v) = \log u - v = -\log(2 - \exp(v)).
\end{equation}
\end{proof}

\paragraph{Squared Hellinger divergence.}

\begin{proposition}
\label{prop:conjugate_squared_hellinger}
The conjugate of $f(u) = (\sqrt{u} - 1)^2$ for $u \in \dom f = [0, +\infty)$ is
$f_*(v) = \frac{v}{1-v}$ for $v \in \dom f_* = (-\infty, 1)$.
\end{proposition}
\begin{proof}
We have $uv - f(u) = u(v-1) + 2 \sqrt{u} -1$.
Hence, for $v \geq 1$, $u \rightarrow uv - f(u)$ is unbounded above.
For $v< 1$, $\lim_{u\rightarrow + \infty} \{uv -f(u)\} = -\infty$ and $uv -f(u)$ 
has then a maximizer $u^\star$ in $[0, +\infty)$ characterized by $f'(u^\star) = v$.
We have 
$f(u) = (\sqrt{u} - 1)^2 = u - 2 \sqrt{u} + 1$.
Using
\begin{align}
v = f'(u) = 1 - \frac{1}{\sqrt{u}}  
\iff \sqrt{u} = \frac{1}{1-v}
\iff u = \frac{1}{(1-v)^2},
\end{align}
we get
\begin{align}
uv &= \frac{v}{(1-v)^2} \\
\sqrt{u} - 1 &= \frac{v}{1-v} \\
(\sqrt{u} - 1)^2 &= \frac{v^2}{(1-v)^2}
\end{align}
so that
\begin{equation}
f^*(v) 
= \frac{v}{(1-v)^2} - \frac{v^2}{(1-v)^2}
= \frac{v}{1-v}.
\end{equation}
\end{proof}

\paragraph{Chi-squared divergence (Pearson divergence).}

\begin{proposition}
\label{prop:conjugate_chi_square}
The conjugate of $f(u) = \frac{1}{2}(u^2 - 1)$ for $u\in \dom f = \RR$ is
$f_*(v) = \frac{1}{2}(v^2 + 1)$ for $v \in \dom f_* = \RR$.
\end{proposition}
\begin{proof}
$f$ is a strongly convex function on $\RR$.
Hence the maximizer $u^\star$ of $u \mapsto uv - f(u)$ exists for any $v \in \RR$ 
and is characterized by $f'(u^\star) = v$.
Using $f'(u) = u$, we get $v = f'(u) = u$.
Therefore,
\begin{equation}
f_*(v) = uv - f(u) = v^2 - \frac{1}{2}(v^2 - 1) = \frac{1}{2}(v^2 + 1).
\end{equation}
\end{proof}

\paragraph{Reverse Chi-squared divergence (Neyman divergence).}

\begin{proposition}
\label{prop:conjugate_reverse_chi_square}
The conjugate of $f(u) = \frac{1}{2}(\frac{1}{u} - 1)$ for $u \in \dom f = (0, +\infty)$ 
is $f_*(v) = -\sqrt{-2v} + \frac{1}{2}$ for $v \in \dom f_* = (-\infty, 0)$.
\end{proposition}
\begin{proof}
For $v > 0$, $\lim_{u \rightarrow + \infty} \{uv - f(u)\} = +\infty$.
For $v <0$, $\lim_{u\rightarrow 0} \{uv - f(u)\} = \lim_{u\rightarrow + \infty} \{uv - f(u)\} = -\infty$. 
Hence the maximizer $u^\star$ of $u\mapsto uv - f(u)$ exists and is characterized by $f'(u^\star) = v$.
Using $v = f'(u) = -\frac{1}{2u^2}$, we get
\begin{equation}
u^2 = -\frac{1}{2v} \iff u = \frac{1}{\sqrt{-2v}}.
\end{equation}
Therefore,
\begin{equation}
f_*(v) 
= uv - \frac{1}{2}\left(\frac{1}{u} - 1\right)
= \frac{v}{\sqrt{-2v}} - \frac{\sqrt{-2v}}{2} + \frac{1}{2} = -\sqrt{-2v} + \frac{1}{2}
\end{equation}
\end{proof}

\paragraph{$\alpha$-divergences.}


\begin{proposition}
\label{prop:conjugate_alpha_div}
The conjugate of
$f(u) = \frac{(u^\alpha - 1) - \alpha(u - 1)}{\alpha(\alpha-1)} + \iota_{\RR_+}(u)$
is
\[
f_*(v) = \begin{cases}
\frac{(1+(\alpha-1)v)^{\frac{\alpha}{\alpha-1}} - 1}{\alpha} &\mbox{ if } \alpha \in (0, 1) \\
\exp(v) &\mbox{ if } \alpha = 1 \\
\frac{(1+(\alpha-1)v)_+^{\frac{\alpha}{\alpha-1}} - 1}{\alpha} & \mbox{ if } \alpha \in (1, +\infty), \ \mbox{where} \ x_+ = \max(x, 0)
\end{cases}
\]
for 
\[
v \in \dom f^* = \begin{cases}
(-\infty, -\frac{1}{\alpha-1}] & \mbox{ if } \alpha \in (0, 1) \\
\RR & \mbox{ if } \alpha \geq 1
\end{cases}
\]
\end{proposition}
\begin{proof}
\textbf{Case $\alpha \in (0,1)$}. 
As $u \rightarrow + \infty$, we have 
$vu - \frac{(u^\alpha - 1) - \alpha(u - 1)}{\alpha(\alpha-1)} \sim \left(v + \frac{1}{\alpha-1}\right)u$.

Hence for $v>\frac{1}{1-\alpha}$, 
$\lim_{u \rightarrow + \infty} vu - \frac{(u^\alpha - 1) - \alpha(u - 1)}{\alpha(\alpha-1)}=+\infty$.

For $v \leq \frac{1}{1-\alpha}$, 
$\lim_{u \rightarrow +\infty} \{ vu - \frac{(u^\alpha - 1) - \alpha(u - 1)}{\alpha(\alpha-1)}\} 
= \lim_{u \rightarrow 0 } \{ vu - \frac{(u^\alpha - 1) - \alpha(u - 1)}{\alpha(\alpha-1)}\} = -\infty$.
Hence the maximizer $u^\star$ of $u\mapsto uv - f(u)$ exists and is characterized by $f'(u^\star) = v$.

We have 
\begin{align*}
    v = f'(u) = \frac{u^{\alpha -1} -1}{\alpha -1} 
    \iff u = (1+(\alpha-1)v)^{\frac{1}{\alpha-1}} = f_*'(v)
\end{align*}
Therefore, denoting $z= (1+ (\alpha-1)v)$, we have, for $v \leq \frac{1}{1-\alpha}$
\begin{align*}
    f^*(v) & = uv - f(u) \\
    & = v z^{\frac{1}{\alpha-1}}- \frac{(z^{\frac{\alpha}{\alpha-1}} -1) - \alpha(z^{\frac{1}{\alpha-1}} -1)}{\alpha (\alpha-1)} \\
    & = \frac{\alpha ((\alpha-1) v + 1) z^{\frac{1}{\alpha-1}} -z^{\frac{\alpha}{\alpha-1}} - (\alpha - 1)}{\alpha(\alpha-1)} \\
    & = \frac{z^{\frac{\alpha}{\alpha -1}} - 1}{\alpha} \\
    & = \frac{(1+(\alpha-1)v)^{\frac{\alpha}{\alpha-1}} - 1}{\alpha}.
\end{align*}

\textbf{Case $\alpha \in (1, +\infty)$}. 
For any $v \in \RR$, $\lim_{u \rightarrow + \infty} vu - \frac{(u^\alpha - 1) - \alpha(u - 1)}{\alpha(\alpha-1)} = -\infty$.
Denote $g(u) = vu - \frac{(u^\alpha - 1) - \alpha(u - 1)}{\alpha(\alpha-1)}$.
Note that $g'$ is decreasing on $[0, +\infty]$.

If $v \leq \frac{1}{1-\alpha}$, we have 
$g'(0) = v + \frac{1}{\alpha-1} \leq 0$, hence $g'(u)<0$ for any $u>0$, that is $g$ is decreasing on $[0, +\infty)$ and
$\sup_{u\geq 0} g(u) = g(0) = -f(0)$ in this case.

If $v > \frac{1}{1-\alpha}$, $g'(0) > 0$, $\lim_{u\rightarrow + \infty} g'(u) = - \infty$ 
and a maximizer $u^\star$ of $g$ can be found 
by the necessary condition $f'(u^\star) = v$. 
Following the computations done in the case $\alpha \in (0, 1)$, for $v > \frac{1}{1-\alpha}$ we get
\begin{align*}
    f^*(v) & = \frac{(1+(\alpha-1)v)^{\frac{\alpha}{\alpha-1}}}{\alpha} \\
    f_*'(v) &  = (1+(\alpha-1)v)^{\frac{1}{\alpha-1}}.
\end{align*}
And for any $v \in \RR$, we get 
\begin{align*}
    f^*(v) & = \frac{(1 +(\alpha-1) v)_+^\frac{\alpha}{\alpha-1} -1}{\alpha} \\
    f_*'(v) & = (1+(\alpha-1)v)_+^{\frac{1}{\alpha-1}}.
\end{align*}
where $x_+ = \max\{x, 0\}$.

\textbf{Case $\alpha=1$} Follows similar computations as for the KL case.
\end{proof}

\subsection{$f$-softargmaxes}

\paragraph{KL divergence.}

With $f_*(v) = f_*'(v) = \exp(v - 1)$, we obtain
\begin{equation}
\tau^\star + 1 = \log \sum_{j=1}^k q_j \exp(\theta_j).
\end{equation}
Plugging $\tau^\star$ back in \eqref{eq:f_softargmax_sol}, we obtain
\begin{align}
p^\star_j &= \frac{q_j\exp(\theta_j)}{\sum_{j=1}^k q_j\exp(\theta_j)},
\end{align}
Other choices of $f$ do not lead to closed-form expressions.

\subsection{$f$-softplus and $f$-sigmoid}
\label{app:f_sigmoids}

In this section, we derive closed-form expressions for some instances of
$f$-softplus and $f$-sigmoid. In this binary classification setting,
we define $\thetav \coloneqq (0, s)$, $\q = (q_0, q_1)$ and $\p^\star = (1-\pi^\star, \pi^\star)$.
By Proposition~\ref{prop:computation}, we seek for
\[
\tau^\star = \argmin_{\tau \in \RR} 
\tau + q_0 f_*'(-\tau) + q_1 f_*'(s - \tau).
\]
Since $f_*$ differentiable and convex, such a scalar $\tau^\star$ is characterized by 
\begin{align}
    -\tau^\star & \in \dom f_*, \label{eq:bound_rkl1}\\
    s-\tau^\star & \in \dom f_*, \label{eq:bound_rkl2} \\
    q_0 f_*'(-\tau^\star) + q_1 f_*'(s - \tau^\star)
    & = 1. \label{eq:root_rkl}
\end{align}

\paragraph{KL divergence.}

\begin{proposition}
Let $f(u) = u \log u$. Then,
\begin{align}
\sigmoid_f(s; \q) & = \frac{q_1\exp(s)}{q_0 + q_1 \exp(s)}.
\end{align}
\end{proposition}
\begin{proof}
Using \eqref{eq:root_rkl} and $f_*'(v) = \exp(v - 1)$, we obtain 
\begin{equation*}
\tau^\star + 1 = \log(q_0 + q_1 \exp(s)).
\end{equation*}
so that
\begin{align}
\pi^\star &= 
\frac{q_1\exp(s)}{q_0 + q_1 \exp(s)}.
\end{align}
\end{proof}

\paragraph{Reverse KL divergence.}

\begin{proposition}

Let $f(u)=-\log(u)$. Then,
\begin{align*}
    \softplus_f(s; \q) & = \tau^\star- q_0\log(\tau^\star) - q_1\log(\tau^\star-s) - (q_0+q_1)\\
    \sigmoid_f(s; \q) & = \frac{q_1}{\tau^\star - s}
\end{align*}
where 
\[
\tau^\star \coloneqq \frac{1}{2}\left(q_0 +q_1 +s + \sqrt{(q_0+q_1+s)^2 - 4 q_0 s}\right).
\]
\end{proposition}
\begin{proof}
Using $f'_*(v) = -1/v$, the root condition~\eqref{eq:root_rkl} can be written as 
\begin{align}
&q_0 f_*'(-\tau) + q_1 f_*'(s - \tau)= 1 \\
\iff& \frac{q_0}{\tau} + \frac{q_1}{\tau -s} = 1 \\
\iff& \frac{q_0(\tau - s) + q_1 \tau}{\tau(\tau-s)} = 1 \\
\iff& a\tau^2 + b\tau + c = 0,
\end{align}
for
\begin{align}
a &\coloneqq 1 \\
b &\coloneqq -(q_0 + q_1 + s) \\
c &\coloneqq q_0 s.
\end{align}
Let us define the discrimant 
\begin{align*}
\Delta & \coloneqq b^2 - 4ac \\
& = (q_0 + q_1 + s)^2 - 4 q_0 s \\
& = (q_1 - q_0 + s)^2 + 4q_0q_1 \\
& > 0.
\end{align*}
Therefore, the root condition~\eqref{eq:root_rkl} reads
\begin{align*}
\tau^\star & \in \{\tau_1, \tau_2\} \\
& \coloneqq \left\{\frac{-b + \sqrt{\Delta}}{2a}, \frac{-b - \sqrt{\Delta}}{2a}\right\}.
\end{align*}

Since $\dom f_* = \RR_-$, we have
\begin{align*}
    (-\tau^\star & \in \dom f_*) 
    \ \mbox{and} \
    (s-\tau^\star \in \dom f_*) \iff  \tau^\star \geq \max\{0, s\}.
\end{align*}
We have
\begin{align*}
    \tau_2 & = \frac{1}{2}\left(q_0 +q_1 +s - \sqrt{(q_0+q_1+s)^2 - 4 q_0 s}\right) \\
    \tau_2 - s & 
    = \frac{1}{2}\left(q_0 + q_1 - s
    -\sqrt{(q_0+q_1-s)^2 + 4 q_1 s}\right).
\end{align*}
Hence for $s<0$, $\tau_2 < 0$, and for $s\geq0$, $\tau_2-s<0$. So the unique solution to the set of conditions~\eqref{eq:bound_rkl1},~\eqref{eq:bound_rkl2} and~\eqref{eq:root_rkl} is
\[
\tau^\star = \tau_1 = \frac{1}{2}\left(q_0 +q_1 +s + \sqrt{(q_0+q_1+s)^2 - 4 q_0 s}\right)
\]
Plugging this value in the formulas for the softplus and the sigmoid in Proposition~\ref{prop:computation} gives the result.
\end{proof}

\underline{Numerically stable implementation.}

For very large positive or negative $s$ we may get numerical issues of the form $as -bs$.
For a numerically stable implementation, one could write
\begin{align*}
  \tau^\star & 
  = q_0 + h(g(s)) \\
  \tau^\star - s &
  = q_1 + h(-g(s))
\end{align*}
where 
\begin{align*}
    g(s) & \coloneqq q_1 - q_0 +s\\
    h(g) & \coloneqq \frac{1}{2} \cdot \begin{cases}
    |g|\left(\sqrt{1 +\frac{4q_0q_1}{g^2}}-1\right) & \mbox{if} \ g < - 1 \\
    g + \sqrt{g^2 + 4q_0q_1}& \mbox{otherwise}
    \end{cases}
\end{align*}
We can then further rewrite
\begin{align*}
    |g|\left(\sqrt{1 +\frac{4q_0q_1}{g^2}}-1\right)
    & = |g| \frac{4q_0q_1/g^2}{\sqrt{1 +\frac{4q_0q_1}{g^2}}+1} \\
    & = \frac{4q_0q_1}{\sqrt{g^2 + 4q_0q_1} + |g|}.
\end{align*}

\paragraph{Jensen-Shannon divergence.}

\begin{proposition}
Let $f(u) = u \log u - (u+1) \log\left(\frac{u+1}{2}\right)$.
Then,
\begin{align}
\softplus_f(s; \q) & = \log(x^\star) - q_0\log(2 - 1/x^\star) - q_1\log(2-1/(x^\star \cdot y)) \\
\sigmoid_f(s; \q) & = \frac{q_1}{2 \cdot x^\star \cdot y - 1},
\end{align}
where
\begin{align}
x^\star &\coloneqq \frac{-b + \sqrt{\Delta}}{2a} \\
y &\coloneqq \exp(-s) \\
a &\coloneqq 4y \\
b &\coloneqq -2(1 + y + y q_0 + q_1)  \\
c &\coloneqq 1 + q_0 + q_1 \\
\Delta &\coloneqq b^2 - 4ac.
\end{align}
\end{proposition}
\begin{proof}
Using $f_*'(v) = \frac{1}{2 \exp(-v) - 1}$, the root equation \eqref{eq:root_rkl}
\begin{equation}
\frac{q_0}{2 \exp(\tau) - 1}
+
\frac{q_1}{2 \exp(\tau - s) - 1} = 1.
\end{equation}
Using the change of variables 
$x \coloneqq \exp(\tau)$ 
and 
$y \coloneqq \exp(-s)$,
we obtain
\begin{align}
&\frac{q_0}{2x - 1} + \frac{q_1}{2xy - 1} = 1 \\
\iff& q_0(2xy - 1) + q_1(2x-1) = (2x-1)(2xy-1) \\
\iff&a x^2 + b x + c = 0,
\end{align}
where we defined
\begin{align}
a &\coloneqq 4y \\
b &\coloneqq -2(1 + y + y q_0 + q_1)  \\
c &\coloneqq 1 + q_0 + q_1.
\end{align}
Let us define the discriminant
\begin{equation}
\Delta \coloneqq b^2 - 4ac.
\end{equation}
We have
\begin{align*}
\Delta
& = 4y\left[\left((1+q_0)y^{1/2} + (1+q_1)y^{-1/2}\right)^2 - 4(1+ q_0+q_1)\right] \\
& \geq 4y\left[ 4(1+q_0)(1+q_1) -  4(1+ q_0+q_1)\right] \\
& = 16y q_0q_1 \\
& > 0,
\end{align*}
where in the second line we used that $y> 0$ and $\min_{x>0} \alpha x + \beta x^{-1} = 2 \sqrt{\alpha\beta}$ for any $\alpha>0, \beta>0$.
Therefore, we have
\begin{equation}
x^\star = \exp(\tau^\star) = \frac{-b + \sqrt{\Delta}}{2a} \\.
\end{equation}
Using \eqref{eq:f_softmax_sol} with
\begin{align}
f_*(-\tau^\star) &= -\log(2 - \exp(-\tau^\star)) = -\log(2 - 1/x^\star) \\
f_*(s-\tau^\star) &= -\log(2 - \exp(s-\tau^\star)) = -\log(2 - 1/(x^\star \cdot y)),
\end{align}
we obtain
\begin{equation}
\softplus_f(s) = \log(x^\star) - q_0\log(2 - 1/x^\star) - q_1\log(2-1/(x^\star \cdot y)).
\end{equation}
Similarly, using \eqref{eq:f_softargmax_sol}, we obtain
\begin{equation}
\sigmoid_f(s)  = \frac{q_1}{2\exp(\tau^\star - s) - 1} = \frac{q_1}{2 \cdot x^\star \cdot y -1}.
\end{equation}
\end{proof}

\underline{Numerically stable implementation.}

Working with exponential requires careful handling of any formula like $uy - vy$ or $(uy)/(vy)$ 
that can easily reduce numerically to $\infty - \infty=\mathrm{NaN}$ or $\infty/\infty = \mathrm{NaN}$.
We therefore derive the expressions of $x^\star$ and $x^\star y$ in terms of numerically stable operations.

We have, denoting $\alpha \coloneqq 1 + q_0$ and $\beta \coloneqq 1 + q_1$,
\begin{align*}
x^\star
& = \frac{\alpha + \beta y^{-1}}{4} + \frac{1}{4}\sqrt{(\alpha + \beta y^{-1})^2 - 4cy^{-1}}
\\
& = \frac{y^{-1/2}}{4}\left(\alpha y^{1/2} + \beta y^{-1/2} + \sqrt{(\alpha y^{1/2} + \beta y^{-1/2})^2 - 4 c}\right) \\
& = \frac{y^{-1/2}}{4}(\alpha y^{1/2} + \beta y^{-1/2})
\left(1+ \sqrt{1- \frac{4(1+q_0+q_1)}{(\alpha y^{1/2} + \beta y^{-1/2})^2}}\right) \\
& = \exp(s/2 + h(s/2)), 
\end{align*}
where
\begin{align*}
    h(s)
    & \coloneqq g(s) + \log\left(1+\sqrt{1 - \frac{4(1+q_0+q_1)}{\exp(2g(s))}} \right) - 2\log(2) \\
    g(s) & \coloneqq \log\left( \alpha e^{-s} + \beta e^s\right) \\
    & = \mathrm{logsumexp}(-s + \log(1+ q_0), s + \log(1 +q_1)).
\end{align*}
To summarize, we have
\begin{align*}
    x^\star & = \exp(s/2 + h(s/2)) \\
    x^\star y &= \exp(-s/2 + h(s/2))
\end{align*}
and therefore
\begin{align*}
\softplus_f(s) & = s/2 + h(s/2) -q_0\log\left[2-\exp(-s/2-h(s/2))\right] - q_1\log\left[2-\exp(s/2-h(s/2))\right] \\
\sigmoid_f(s) &  = \frac{q_1}{2\exp(-s/2 + h(s/2)) - 1}.
\end{align*}

\paragraph{Squared Hellinger divergence.}

Using $f_*'(v) = \frac{1}{(1-v)^2}$
and using the change of variable $x \coloneqq \tau + 1$, the root equation~\eqref{eq:root_rkl} becomes
\begin{align}
\frac{q_0}{(1+\tau)^2} + \frac{q_1}{(1+\tau-s)^2} = 1
&\iff \frac{q_0}{x^2} + \frac{q_1}{(x-s)^2} = 1 \\
&\iff q_0 (x-s)^2 + q_1 x^2 = x^2(x-s)^2 \\
&\iff q_0 (x^2 - 2sx + s^2) + q_1 x^2 = x^2(x^2 - 2sx + s^2) \\
&\iff q_0 (x^2 - 2sx + s^2) + q_1 x^2 = x^4 - 2sx^3 + s^2 x^2 \\
&\iff ax^4 + bx^3 + cx^2 + dx + e = 0,
\end{align}
where
\begin{align}
a &\coloneqq 1\\
b &\coloneqq -2s \\
c &\coloneqq s^2 - q_0 - q_1\\
d &\coloneqq 2s q_0 \\
e &\coloneqq -q_0 s^2.
\end{align}
This is a quartic equation, which can be solved in closed form.





\subsection{Proof of Proposition \ref{prop:computation}}
\label{proof:computation}

First, the maximum defining the softmax and softargmax is well defined since it is a strictly concave problem on a non-empty bounded set.
Denoting $\Omega_j(p) \coloneqq q_j f(p/q_j)$, such that $\Omega_j'(p) = f'(p/q_j)$ and $\Omega_j^*(\theta) = q_j f^*(\theta)$,
we can apply Lemma~\ref{prop:dual_separable_simplex} and get
\begin{align}
    \max_{\p \in \triangle^k} \langle \p, \thetav \rangle - \sum_{j=1}^k q_j f(p_j/q_j)
    & = \inf_{\tau \in \RR} \tau + \sum_{j=1}^k q_j f^*(\max\{\theta_j-\tau, f'(0)\}),
    \label{eq:softmax_dual}
\end{align}
with $f'(0) \coloneqq \lim_{x \rightarrow 0, x \geq 0} f'(x) \in \RR \cup \{-\infty\}$.

Since $(0, +\infty) \subseteq \dom f'$, and $\q >0$, $f'\left(\left(\sum_{j=1}^k q_j\right)^{-1}\right)$ and $f'(q_{j_{\max}}^{-1})$ for $j_{\max} \in \argmax_{j\in \{1, \ldots, k\}} \theta_j$
are well defined. We can then define
\begin{align*}
    \tau_{\min} & \coloneqq \theta_{\max} - f'(q_{j_{\max}}^{-1}) \\
    \tau_{\max} & \coloneqq \theta_{\max} - f'\left(\left(\textstyle{\sum_{j=1}^k} q_j\right)^{-1}\right)
\end{align*}
where $\theta_{\max} = \theta_{j_{\max}}$. Since $\q>0$, $q_{j_{\max}} <\sum_{j=1}^k q_j$, and since $f'$ is increasing, we have $\tau_{\min} < \tau_{\max}$.

We can then analyze the following function on $[\tau_{\min}, \tau_{\max}]$:
\begin{align*}
    h(\tau) & \coloneqq \tau + \sum_{j=1}^k q_j f^*(\max\{\theta_j-\tau, f'(0)\}).
\end{align*}

First, we need to ensure that we can compute derivatives of this function in $[\tau_{\min}, \tau_{\max}]$.
From Lemma~\ref{prop:conj_sum_cvx_pos}, we have $ f^*(\max\{\theta_j-\tau, f'(0)\}) = (f+\iota_{\RR_+})^*(\theta_j -\tau)$ and $\dom (f+\iota_{\RR_+})_*' = \dom f_*' \cup (-\infty, f'(0)]$. 
Therefore, if $f'(0) > -\infty$, the domain of $(f+\iota_{\RR_+})_*'$ is unbounded below.
Otherwise, if $f'(0) = -\infty$, since $\im f' \subseteq \dom f_*'$,
the domain of $f_*'$, and so of $(f+\iota_{\RR_+})_*'$, are unbounded below.
Denoting $\alpha \coloneqq \sup \dom (f+\iota_{\RR_+})_*'$, 
since $\theta_{\max} - \tau_{\min} = f'(q_{j_{\max}}^{-1}) \in \dom f_*'$, 
we then have  $\theta_{\max} - \tau_{\min} < \alpha$ and therefore
\begin{align*}
    & \tau \geq \tau_{\min} \\
    \implies & \theta_{\max} - \tau < \alpha \\
    \iff & \theta_j - \tau < \alpha, \mbox{for all} \ j \in \{1, \ldots, k\} \\
    \implies & \tau \in \dom h'.
\end{align*}
We can then show that $h'(\tau_{\min}) \leq 0$ and $h'(\tau_{\max}) \geq 0$. Indeed, we have
\begin{align*}
    h'(\tau_{\min}) 
    & = 1 - \sum_{j=1}^k q_j f_*'(\max\{\theta_j - \tau_{\min}, f'(0)\}) \\
    & \stackrel{(i)}{\leq} 
    1 - q_{j_{\max}} f_*'(\max\{\theta_{\max} -\tau_{\min}, f'(0)\}) \\
    & = 1 - q_{j_{\max}}f_*'(\max\{f'(q_{j_{\max}}^{-1}), f'(0)\}) \\
    & \stackrel{(ii)}{=} 0,
\end{align*}
where in $(i)$ we used that $\q>0$ and $f_*'(\max\{y, f'(0)\}) \geq 0$ for any $y \in \dom (f + \iota_{\RR_+})_*'$ as per Lemma~\ref{prop:conj_sum_cvx_pos}, and in $(ii)$, we used that $q_{j_{\max}}^{-1} > 0$, $f'$ is increasing and $f_*'(f'(p)) = p$ for any $p \in \dom f'$.

Similarly, we have
\begin{align*}
    h'(\tau_{\max}) 
    & = 1 - \textstyle{\sum_{j=1}^k} q_j f_*'(\max\{\theta_j - \tau_{\max}, f'(0)\}) \\
    & \stackrel{(i)}{\geq} 
    1 - \left(\textstyle{\sum_{j=1}^k} q_j\right) f_*'(\max\{\theta_{\max} -\tau_{\max}, f'(0)\}) \\
    & = 1 -\left(\textstyle{\sum_{j=1}^k} q_j\right)
    f_*'\left(\max\left\{f'\left(\left(
    \textstyle{\sum_{j=1}^k} q_j\right)^{-1}\right), 
    f'(0)\right\}\right) \\
    & \stackrel{(ii)}{=} 0,
\end{align*}
where in $(i)$ we used that $\sum_j a_j b_j \leq (\sum_j a_j) \max_j b_j$ 
if $a_j \geq 0$ with here $a_j = q_j > 0$ and $b_j = f_*'(\max\{\theta_j - \tau_{\max}, f'(0)\})$, 
and in $(ii)$ we used the same reasoning as for $h'(\tau_{\min})$.

Finally, we show that $h'$ is increasing on $[\tau_{\min}, \tau_{\max}]$.
For $j = j_{\max}$, we have that 
\begin{align*}
    & \tau \leq \tau_{\max} \\
  \iff  & \theta_{\max} - \tau \geq \theta_{\max} - \tau_{\max} = f'(q_{j_{\max}}) > f'(0).
\end{align*}
Since $f_*'(\max\{\theta, f'(0)\}) = (f+ \iota_{\RR_+})^*$ is increasing on $\dom f^* \setminus (-\infty, f(0)]$ (see Proposition~\ref{prop:conj_sum_cvx_pos}), $\tau \mapsto - f_*'(\max\{\theta_{\max} - \tau, f'(0)\})$ is increasing on $[\tau_{\min}, \tau_{\max}]$ and so $h'(\tau) = 1 - \sum_{j=1}^k q_j f_*'(\max\{\theta_j - \tau, f'(0)\})$ is increasing on $[\tau_{\min}, \tau_{\max}]$ (as a sum of an increasing and non decreasing function).


Overall, $h$ is well defined and strictly convex on $[\tau_{\min}, \tau_{\max}]$ such that $h'(\tau_{\min}) \leq 0$ and $h'(\tau_{\max}) \geq 0$. Hence, we have 
\[
\inf_{\tau \in \RR} h(\tau) = \min_{\tau_{\min} \leq \tau \leq \tau_{\max}} h(\tau).
\]
and the unique minimizer $\tau^\star$ can then be found by solving the first order optimality condition $h'(\tau^\star) = 0$ in $[\tau_{\min}, \tau_{\max}]$. This gives the expression of $\tau^\star$ in the claim and plugging $\tau^\star$ back into~\eqref{eq:softmax_dual} gives the expression of the softmax. The expression of the softargmax follows from Lemma~\ref{prop:dual_separable_simplex}.

\subsection{Lemmas}

\begin{lemma}\label{prop:dual_separable_simplex}
Given $k$ strictly convex differentiable univariate scalar functions $\Omega_j$ such that $(0, +\infty) \subseteq \dom \Omega_j$,
for any $\thetav \in \RR^k$, we have
\begin{align*}
    \sup_{\p \in \triangle^k} \langle \p, \thetav\rangle - \sum_{j=1}^k \Omega_j(p_j) 
    = \inf_{\tau \in \RR} \tau + \sum_{j=1}^k (\Omega_j + \iota_{\RR_+})^*(\theta_j-\tau),
\end{align*}
where
\[
(\Omega_j + \iota_{\RR_+})^*(z) = 
\Omega_j^*(\max\{\theta_j-\tau, \Omega_j'(0)\}) 
\]
and $\Omega_j'(0) \coloneqq \lim_{x\rightarrow 0, x\geq 0} \Omega_j'(x) \in \RR \cup \{-\infty\}$.
Given a minimizer $\tau^\star$ of the right hand-side, the maximizer $\p^\star$ of the left hand side is given by 
\begin{align*}
    p_j^\star = (\Omega_j + \iota_{\RR_+})_*'(\theta_j-\tau^\star)
    = (\Omega_j^*)'(\max\{\theta_j - \tau^\star, \Omega_j'(0)\}).
\end{align*}
\end{lemma}
\begin{proof}
We have
\begin{align*}
    \sup_{\p \in \triangle^k} \langle \p, \thetav\rangle - \sum_{j=1}^k \Omega_j(p_j)
    & = \sup_{\p \in \RR_+^k} \inf_{\tau \in \RR}  
    \langle \p, \thetav\rangle - \sum_{j=1}^k \Omega_j(p_j) + (1- \langle \p, \ones\rangle)\tau \\
    & = \inf_{\tau \in \RR} \tau + \max_{\p \in \RR_+^k}
    \langle \p, \thetav - \tau \ones\rangle - \sum_{j=1}^k \Omega_j(p_j) \\
    & = \inf_{\tau \in \RR} \tau + \sum_{j=1}^k (\Omega_j +\iota_{\RR_+})^*(\theta_j-\tau),
\end{align*}
where the second equality stands from strong duality, using that the simplex is convex, 
that the $\Omega_j$ are convex functions, and that
the maximization problem is strictly feasible since $(0, + \infty) \subseteq \dom \Omega_j$.

We can then apply Lemma~\ref{prop:conj_sum_cvx_pos}, and get
\begin{align*}
    \max_{\p \in \triangle^k} \langle \p, \thetav\rangle - \sum_{j=1}^k \Omega_j(p_j)
    & = \inf_{\tau \in \RR} \tau + \sum_{j=1}^k \Omega_j^*(\max\{\theta_j-\tau, \Omega_j'(0)\}),
\end{align*}
with $\Omega_j'(0) = \lim_{x\rightarrow 0,  x\geq 0} \Omega_j'(x) \in \RR \cup \{-\infty\}$.
Given a minimizer $\tau^\star$ of the right hand-side, the maximizer $\p^\star$ 
of the left hand side is given by strong duality as
\begin{align*}
    p_j^\star = \argmax_{p_j\geq 0} p_j(\theta_j-\tau) -\Omega_j(p_j), 
\end{align*}
whose full expression follows from Lemma~\ref{prop:conj_sum_cvx_pos}.
\end{proof}

\begin{lemma}\label{prop:conj_sum_cvx_pos}
Given a strictly convex differentiable univariate scalar function $f$ such that 
$(0, +\infty) \subseteq \dom f$,
we have, for any $y \in \RR$,
\begin{align*}
    (f + \iota_{\RR_+})^*(y) \coloneqq \sup_{x\geq 0} xy - f(x) 
    = f^*(\max\{y, f'(0)\})
\end{align*}
with $f'(0) \coloneqq \lim_{x\rightarrow 0, x\geq 0} f'(x) \in \RR \cup \{-\infty\}$.
For 
$y \in \dom (f + \iota_{\RR_+})_*'
= \dom f_*' \cup (-\infty, f'(0)]$,
we have
\begin{align*}
    (f + \iota_{\RR_+})_*'(y)
    = f_*'(\max\{y, f'(0)\}) \geq 0.
\end{align*}
Finally, $(f+\iota_{\RR_+})_*'$ is increasing on $(\dom f + \iota_{\RR_+})_*' \setminus (-\infty, f'(0)]$.
\end{lemma}
\begin{proof}
Fix some $y\in \RR$, denote $f'(0) = \lim_{x\rightarrow 0, x \geq 0} f'(x) \in \RR \cup\{-\infty\}$ and 
\[
h(x) \coloneqq xy - f(x).
\]
If $y \leq f'(0)$ (so provided that $f'(0) > -\infty$), then $h'(x) = y-f'(x) > 0$ for all $x > 0$, 
since $f'$ is increasing. So 
$h$ is decreasing on $(0, +\infty)$ and the maximum of
$h$ on $\RR_+$ is reached at $0$, that is 
\[
\sup_{x\geq 0} xy - f(x) = -f(0).
\]
If $y \geq f'(0)$, then $h'(x) = y - f'(x) > 0$ for all $x < 0$ since $f'$ is increasing.
Therefore $h(x) < h(0)$ for all $x < 0$. Finally, since $h'(0) \leq 0$, the supremum of $h$ on $\RR$
is necessarily greater or equal than $0$, that is, 
\[
\sup_{x \geq 0} xy - f(x) = \sup_{x\in \RR} xy -f(x) = f^*(y).
\]
Note that for $y = f'(0)$ (provided that $f'(0) > -\infty$), we then get
\[
\sup_{x\geq 0} x f'(0) - f(x) = - f(0) = f^*(f'(0)),
\]
and 
\[
f'_*(f'(0)) = \argmax_{x \in \RR} \{xf'(0) - f(x)\} = 0.
\]
Combining the two cases above, we get
\begin{align*}
\sup_{x\geq 0} xy - f(x) 
& = \begin{cases}
f^*(y) & \mbox{if} \ y > f'(0) \\
-f(0) & \mbox{if} \ y \leq f'(0)
\end{cases} \\
& = f^*(\max\{y, f'(0)\}).
\end{align*}
For the derivative, first note that since $f + \iota_{\RR_+}$ is strictly convex, 
its convex conjugate is differentiable on its domain of definition.
Then, given $y \in \dom(f + \iota_{\RR_+})_*'$, we have from the previous considerations,
\begin{align*}
    (f + \iota_{\RR_+})_*'(y) 
    = \argmax_{x\geq 0} \{xy - f(x)\}
    = \begin{cases}
    f_*'(y) & \mbox{if} \ y > f'(0) \\
    0 & \mbox{if} \ y \leq f'(0)
    \end{cases}.
\end{align*}
Using that $f_*'(f'(0)) = 0$,
we get 
\[
(f + \iota_{\RR_+})_*'(y) = f_*'(\max\{y, f'(0)\}).
\]
The expressions above also show that $\dom(f + \iota_{\RR_+})_*' = \dom f_*' \cup (-\infty, f'(0)]$.

In any case, for $y \in \dom ((f+\iota_{\RR_+})^*)'$,
\[
(f + \iota_{\RR_+})_*'(y) = \argmax_{x \geq 0} xy - f(x) \geq 0,
\]
by definition of the maximization set.

Finally, since $f$ is strictly convex and differentiable on $(0, +\infty)$, $f'$ is invertible on $(f'(0), f'(\infty))$ 
where $f'(0) = \lim_{x\rightarrow 0, x\geq 0} f'(x)$ and $f'(\infty) = \lim_{x\rightarrow +\infty} f'(x)$. 
Moreover, for any 
$y \in (f'(0), f'(\infty))$,
we then have $(f + \iota_{\RR_+})_*'(y) = f_*'(y) = (f')^{-1}(y)$, 
where $(f')^{-1}$ is the inverse of $f'$ on $(f'(0), f'(\infty))$. Since $f'$ is increasing on $(0, +\infty)$, $(f')^{-1}$ is also increasing on $(f'(0), f'(\infty))$, and so are $f_*'$ and $(f + \iota_{\RR_+})_*'$. Noting that $(f'(0), f'(\infty)) = \dom f_*' \setminus (-\infty, f'(0)] = (\dom f + \iota_{\RR_+})_*' \setminus (-\infty, f'(0)]$ concludes the claim.
\end{proof}



\end{document}